\documentclass[lettersize,journal]{IEEEtran}
\usepackage{amsmath}
\usepackage{amsfonts}
\usepackage{algorithmic}
\usepackage{algorithm}
\usepackage{array}
\usepackage{textcomp}
\usepackage{stfloats}
\usepackage{url}
\usepackage{verbatim}
\usepackage{graphicx}
\usepackage{subfigure}
\usepackage{cite}
\usepackage{amssymb}
\usepackage{enumitem}
\usepackage{color}
\begin{document}

\title{ Riemannian Complex Hermit Positive Definite Convolution Network for Polarimetric SAR Image Classification}

\author{Junfei Shi,~\IEEEmembership{Senior Member, IEEE }, Yuke Li, Mengmeng Nie, Fang Liu,~\IEEEmembership{Senior Member, IEEE }, Haiyan Jin ~\IEEEmembership{Member, IEEE}, Junhuai Li ~\IEEEmembership{Member, IEEE},  Weisi Lin ~\IEEEmembership{Fellow, IEEE} \vspace{-2em}

\thanks{Junfei Shi, Yuke Li, Mengmeng Nie, Haiyan Jin and Junhuai Li were with the Department of Computer Science and Technology, Shaanxi Key Laboratory for Network Computing and Security Technology, Xi'an University of Technology, Xi'an, China. Corresponding author: Weisi Lin. Email: WSLin@ntu.edu.sg.}
\thanks{Fang Liu was with the School
of Artificial Intelligence, Xidian University, Xi'an, China.}
\thanks{Weisi Lin was with the School of Computer Science and Engineering, Nanyang Technological Universsity, Singapore 639798.}
}


\markboth{Journal of \LaTeX\ Class Files,~Vol.~14, No.~8, July~2023}%
{Shell \MakeLowercase{\textit{et al.}}: A Sample Article Using IEEEtran.cls for IEEE Journals}


\maketitle

\begin{abstract}
Deep learning has been extensively utilized for PolSAR image classification. However, most existing methods transform the polarimetric covariance matrix into a real- or complex-valued vector to comply with standard deep learning frameworks in Euclidean space. This approach overlooks the inherent structure of the covariance matrix, which is a complex Hermitian positive definite (HPD) matrix residing in the Riemannian manifold. Vectorization disrupts the matrix structure and misrepresents its geometric properties. To mitigate this drawback, we propose HPDNet, a novel framework that directly processes HPD matrices on the Riemannian manifold. The HPDnet fully considers the complex phase information by decomposing a complex HPD matrix into the real- and imaginary- matrices. The proposed HPDnet consists of several HPD mapping layers and rectifying layers, which can preserve the geometric structure of the data and transform them into a more separable manifold representation. Subsequently, a complex LogEig layer is developed to project the manifold data into a tangent space, ensuring that conventional Euclidean-based deep learning networks can be applied to further extract contextual features for classification. Furthermore, to optimize computational efficiency, we design a fast eigenvalue decomposition method for parallelized matrix processing. Experiments conducted on three real-world PolSAR datasets demonstrate that the proposed method outperforms state-of-the-art approaches, especially in heterogeneous regions.
\end{abstract}


\section{Introduction}

\IEEEPARstart{O}ver the past decades, polarimetric synthetic aperture radar (PolSAR) has garnered significant attention in the field of remote sensing image processing due to its ability to transmit and receive electromagnetic signals under all-weather and all-day conditions. With the rapid advancement of radar imaging technologies, a large volume of high-resolution PolSAR data has been acquired. Thanks to its rich scattering information, PolSAR imagery has been widely utilized in various applications, including image classification \cite{9619948}, agricultural monitoring \cite{10770247}, target recognition \cite{9345705}, and change detection \cite{10456921}. Among various applications, PolSAR image classification has become a fundamental task in image interpretation and has received considerable research interest.

For decades, a variety of classification methods have been developed for PolSAR images, including approaches based on scattering mechanisms \cite{9143461}, target decomposition \cite{10531285}, and statistical distributions \cite{8438543}. These methods typically extract widely used features for PolSAR image analysis, such as Cloude decomposition \cite{10335659}, Freeman decomposition \cite{8602439}, and the Wishart distribution \cite{7018953}. However, traditional approaches alone are insufficient, as they are often sensitive to speckle noise and lack the ability to capture high-level semantic information.

Recently, deep learning methods\cite{10507032,10770247} have been widely used in various fields of remote sensing images, due to their advantages of being the end-to-end learning framework and automatically learning high-level features. Taking into account the polarimetric information, many deep learning models have been proposed for PolSAR image classification, including Convolution Neural Network(CNN)\cite{10685476}, Graph Convolution Network(GCN)\cite{9779320}, Generative attack network (GAN)\cite{9524508}, Transformer\cite{10746331}, etc. However, these methods need to convert the complex covariance matrix into a 9-dimensional real vector to input to the network, which completely ignores the complex phase information. Realizing this shortage, some complex-valued CNN variants\cite{10650936} have been proposed for PolSAR images such as CV-CNN\cite{9323621},CV-3D-CNN\cite{rs9010067}, hybrid CVNet\cite{10693615}, complex couturlet-CNN\cite{10415179} etc. Various deep learning models have been expanded to the complex data domain to learn PolSAR complex scattering characteristics. These methods transform the complex covariance matrix into a 6-dimensional complex vector to fed into the model, effectively learning complex scattering information. Converting a PolSAR complex matrix into a complex vector has been a great advancement for PolSAR deep learning models, which begin to consider the scattering information from the imaginary part. However, the original PolSAR data is a complex covariance matrix, known as the HPD(Hermit Positive Definite) matrix, which is endowed with a Riemannian manifold\cite{7947120}. These existing models are still grounded in Euclidean space, where complex matrices are vectorized, disrupting their intricate structure and overlooking the manifold geometric characteristics inherent in the original PolSAR complex matrices.

In summary, existing deep learning methods for PolSAR image classification within Euclidean space still encounter the following limitations:
\begin{itemize}
\item Conventional deep learning approaches often vectorize the complex scattering matrix, which disrupts its intrinsic geometric structure and inter-channel correlations. Employing Euclidean metrics to process the PolSAR data fails to effectively capture the geometry structure of the covariance matrix, often leading to inaccurate measurements.
\item Conventional Euclidean deep neural networks are not well-suited for directly learning from complex covariance matrices and often fail to capture the intrinsic geometric structure of HPD matrices that reside on Riemannian manifolds. Consequently, they may suffer from feature entanglement and have difficulty extracting discriminative features from PolSAR images.
\end{itemize}
To overcome the aforementioned limitations, it is worth considering whether a deep learning model based on complex HPD matrices can be developed to effectively capture both the complex scattering information and the geometric structure of the matrices.

To learn manifold features from covariance matrices, several Riemannian manifold networks have been developed for natural image processing, demonstrating clear advantages in capturing manifold geometric characteristics. Among these, the SPD network\cite{Huang2017} is the most fundamental work to learn the SPD matrix in Riemannian rather than Euclidean space, which redefines the SPD convolution, ReLu and Pooling operators in manifold space. Building upon SPDNet, several extensions have been proposed to adapt various network architectures from Euclidean to Riemannian space, such as SPD-UNet \cite{wang2023u} and DMT-Net \cite{zhang2020deep}. However, these methods are inherently limited by their focus on SPD matrices and do not account for the complex-valued nature of Hermitian positive-definite (HPD) matrices. Furthermore, they are tailored for natural images and neglect the complex scattering characteristics critical in PolSAR data. Therefore, we should design a new complex HPD network to learn PolSAR data effectively in the complex HPD manifold. Such a network should treat the real and imaginary components with equal importance, as the imaginary part contains critical scattering information, such as the scattering angle, which is vital for distinguishing target objects for PolSAR images.

To address these shortcomings, we propose a novel complex HPD network that can capture the manifold geometry structure of complex matrices and the channel correlation inherent in PolSAR data.
Firstly, we decompose the complex HPD matrix into real and imaginary parts, respectively. Then, an HPD network is designed to convert covariance matrices into a more discriminating manifold space. To learn contextual information in Euclidean space, an HPD LogEig layer is designed to project the complex HPD matrix onto a tangent space, enabling the application of Euclidean operations. After projecting complex HPD matrices to a tangent space, a complex-valued 3DCNN is applied to learn contextual information for classification. In addition, to accelerate computational speed, we design the iterative complex HPD matrix square root normalization method (HPD-ISRT) to optimize matrix eigenvalue decomposition, completing parallel computing.

The main contributions of the proposed complex HPD convolution network can be summarized in three key aspects.

\begin{itemize}
\item[1)] To our knowledge, this is the first time a new complex HPD network is proposed for PolSAR images in Riemannian space. It defines the complex HPD mapping layer, Rectifying layer, and LogEig layer. This network ensures that it can learn geometry characteristics of complex matrices and obtain a more discriminating manifold representation.
\item[2)] A novel Remain-to-Euclidean framework is proposed, which consists of a complex HPDnet in Riemannian space and the CV-3DCNN in Euclidean space. This design project manifold features into Euclidean space, allowing seamless integration with existing deep learning architectures. As a result, the model effectively captures both the geometric structure inherent in Riemannian space and the contextual high-level semantics in Euclidean space.
\item[3)] To reduce the calculation complexity of the eigenvalue decomposition for complex HPD matrix, a revised complex HPD iteration model is defined to accelerate the network and perform parallel conduction on GPU.
\end{itemize}

The remainder of this paper is organized as follows. Section 2 presents the motivation and related work. Section 3 details the proposed methodology. Experimental results and analysis are provided in Section 4. Finally, Section 5 concludes the paper.

\section{ Motivation and Related Work}

\subsection{Motivation}
For the PolSAR image, each data is a complex HPD matrix, which is embodied in a Riemannian manifold. Several manifold metrics have been proposed to measure the geometric distance between two points along a curved surface, such as the affine invariant Riemannian metric (AIRM)\cite{9963700}, Jeffrey and log-Euclidean distances\cite{log2015}, etc. For HPD matrices, the manifold metric can maintain the geometric curve of the HPD matrices, while the Euclidean metric ignores the manifold structure of HPD matrices. So, a Euclidean metric will produce an inaccurate distance and damage the intrinsic geometric relationship of the PolSAR data. In addition, compared to the Euclidean metric (F-norm), manifold metric, such as the log-Euclidean metric(noted by "logE"), has some properties: affine invariance, scale invariance, and complete metric space\cite{minh2017covariances}.

An example is given to show that the manifold metric can give a more accurate measurement than the Euclidean one for two HPD matrices, while the Euclidean metric will amplify the distance of two similar HPD matrices and shorten the distance of different ones.

For two similar HPD matrices A and B in the same class, defined as
\begin{equation}
\small
A = \left( {\begin{array}{*{20}{c}}
   2 & 0  \\
   0 & 2  \\
\end{array}} \right),B = \left( {\begin{array}{*{20}{c}}
   3 & 0  \\
   0 & 3  \\
\end{array}} \right)
\end{equation}

The Euclidean distance after vectorization is calculated as:
\begin{equation}\label{ep1}
\small
 \begin{array}{l}
{d_E}\left( {A,B} \right) = {\left\| {A - B} \right\|_F} = \sqrt 2  \approx 1.414
  \end{array}
\end{equation}

The manifold distance with the Log-Euclidean metric is:
\begin{equation}\label{ep2}
\small
 \begin{array}{l}
 {d_{\log  - E}}\left( {A,B} \right) = {\left\| {\log \left( A \right) - \log \left( B \right)} \right\|_F} \\
  = {\left\| {\left( {\begin{array}{*{20}{c}}
   {\log \left( {\frac{2}{3}} \right)} & 0  \\
   0 & {\log \left( {\frac{2}{3}} \right)}  \\
\end{array}} \right)} \right\|_F} \approx 0.573 \\
 \end{array}
\end{equation}

It can be seen that the log-Euclidean metric has a smaller distance than the Euclidean metric, and the Euclidean metric amplifies the distance between two similar HPD matrices.

However, for two different HPD matrices in the different class, noted by $A$ and $B$ as:
\begin{equation}\label{ep3}
\small
A= \left( {\begin{array}{*{20}{c}}
   1 & 0  \\
   0 & 1  \\
\end{array}} \right),B= \left( {\begin{array}{*{20}{c}}
   1 & {0.9}  \\
   {0.9} & 1  \\
\end{array}} \right)
\end{equation}
where $A$ is a unity matrix, while $B$ has larger non-dialog elements.

Then, the Euclidean distance after vectorization is :
\begin{equation}\label{ep4}
\small
{d_E}\left( {A,B} \right) = {\left\| {A - B} \right\|_F} = \sqrt {{{0.9}^2} + {{0.9}^2}}  \approx 1.273
\end{equation}

While the manifold distance is:
\begin{equation}\label{ep5}
\small
\begin{array}{l}
 {d_{\log  - E}}\left( {A,B} \right) = {\left\| {\log \left( A \right) - \log \left( B \right)} \right\|_F} \\
 {\rm{                  }} = {\left\| {\left( {\begin{array}{*{20}{c}}
   { - 0.641} & 0  \\
   0 & {2.302}  \\
\end{array}} \right)} \right\|_F} \approx 2.39 \\
 \end{array}
\end{equation}

That is, the log-Euclidean distance is much larger than the Euclidean distance for two different matrices, which means that the manifold metric can better measure the HPD data. Based on the properties above, deep networks with Euclidean operations are unsuitable for HPD matrices, whereas more complicated Riemannian networks with manifold metric should be exploited for PolSAR data.

\subsection{Euclidean Deep learning method for PolSAR image classification}

In recent years, deep learning methods have become the cornerstone in the development of Polarimetric Synthetic Aperture Radar (PolSAR) image classification, achieving remarkable performance. A variety of deep learning approaches have been proposed to extract PolSAR features in Euclidean space. However, each resolution cell in PolSAR data is represented by a 3 × 3 complex-valued covariance matrix, which poses a challenge for direct application in standard Euclidean-based deep learning frameworks. To address this, the covariance matrix is typically converted into a 9-dimensional real-valued vector, enabling compatibility with conventional neural network architectures. Commonly used network architectures for PolSAR image classification include Convolutional Neural Networks (CNN)\cite{10538121}, Graph Convolutional Networks (GCN)\cite{9953965,shi2023cnn}, Stacked Autoencoders (SAE)\cite{10201469}, U-Net\cite{wu2023}, and Transformers\cite{10422857}, among others. For instance, Hua et al.\cite{9172110} proposed a three-channel CNN capable of learning features from different polarization channels. Ren et al.\cite{10296522} introduced an incremental learning approach to improve classification performance on unbalanced datasets. Dong et al.\cite{10360854} developed a causal inference-guided feature enhancement framework to more effectively extract discriminative features for PolSAR image classification. Shi et al.\cite{shi2023cnn} proposed a CNN-enhanced fuzzy GCN method to improve classification accuracy at image boundaries. Wang et al.\cite{10401978} designed a superpixel-based multi-scale GCN that captures terrain objects at different scales, thereby enhancing classification performance. Geng et al.\cite{10422857} proposed a hierarchical scattering–spatial interaction Transformer to more effectively capture complex spatial and polarimetric dependencies in PolSAR data.

However, most existing deep learning methods overlook the phase information inherent in radar scattering. To address this limitation, complex-valued convolutional neural network (CV-CNN) models have been introduced for PolSAR image analysis, enabling more effective learning of complex-valued data. Based on this foundation, several complex-valued variants have been proposed to better capture both amplitude and phase characteristics. For instance, Tan et al.\cite{8864110} proposed a complex-valued 3D-CNN model designed to learn features from 3D image blocks. Li et al.\cite{2019Complex} introduced a complex contourlet CNN model that simultaneously captures frequency-domain phase information and spatial texture features. Jiang et al.\cite{jiang2022unsupervised} developed an unsupervised approach to learning complex-valued sparse features, which incorporates population and lifetime sparsity constraints, enabling the extraction of non-redundant features for PolSAR classification. Additionally, Liu et al. \cite{liu2023unified} proposed a unified framework that jointly learns features from multi-polarimetric and dual-frequency SAR data to enhance classification performance. Although complex-valued networks can capture phase information by converting complex matrices into complex-valued vectors, this transformation often disrupts the original matrix structure and inter-channel correlations. As a result, these models struggle to preserve and learn the inherent geometric properties of complex-valued matrices. To address this limitation, it is essential to develop deep learning frameworks that operate directly on complex matrices, enabling more effective representation and classification of PolSAR data.

\subsection{Remaninian Deep learning methods}

Recently, Riemannian deep learning methods tailored to symmetric positive definite (SPD) matrices have gained significant attention, particularly in computer vision. Fiori et al. \cite{fiori2011riemannian} introduced Riemannian gradient-based learning on the complex matrix hypersphere, laying the theoretical foundation for this field. Chakraborty et al. \cite{chakraborty2015recursive} proposed a recursive Fréchet mean computation method on the Grassmannian manifold, which has also been applied to vision tasks. Building on these foundations, Huang et al. \cite{huang2017riemannian} developed the first Riemannian network for SPD matrix learning and later extended it to deep learning on Lie groups for skeleton-based action recognition. Additionally, Chakraborty et al. \cite{chakraborty2018statistical} introduced a statistical recurrent model for processing SPD matrices. Subsequently, various deep learning methods have been developed on SPD matrices within Riemannian spaces. Zhang et al.\cite{zhang2020deep} proposed a deep manifold-to-manifold transformation network for action recognition. Wang et al.\cite{wang2021symnet} introduced SymNet, a lightweight SPD manifold learning framework for image classification. Sukthanker et al.\cite{sukthanker2021neural} explored neural architecture search on SPD manifolds. Wang et al.\cite{wang2020multiple} developed a multi-kernel metric learning approach to construct manifold-valued descriptors. Chakraborty et al. \cite{chakraborty2020manifoldnet} presented ManifoldNet, demonstrating its effectiveness in computer vision tasks. Most recently, Chen et al. \cite{chen2023riemannian} proposed an optimization strategy to accelerate matrix computations in Riemannian manifold networks. Manifold learning methods based on symmetric positive definite (SPD) matrices have demonstrated the ability to capture the geometric structure of SPD data, providing improved representation for complex manifold data. However, these methods are not fully suitable for Polarimetric Synthetic Aperture Radar (PolSAR) data, which is represented by Hermitian positive definite (HPD) matrices. Unlike SPD matrices, HPD matrices consist of both real and imaginary components, making their structure more complex and computationally demanding within deep networks. Moreover, the imaginary part carries critical phase information that cannot be neglected. Therefore, it is essential to develop a novel HPD-based manifold learning framework specifically tailored for classification of PolSAR images.

\section{Proposed method}
In this paper, we develop a Riemannian complex HPD convolution network for polarimetric SAR image classification, noted by "HPD\_CNN". The proposed method consists of two modules: the Riemannian complex HPD network and the CV-3DCNN enhanced network. Firstly, to learn HPD matrix, the HPD covariance matrix unfolds as the addition of real-part and imaginary-part matrices. Then, a complex HPD network is designed by defining the HPD mapping layer, nonlinear HPD Rectifying layer and complex HPD LogEig layer. Thus, the HPD matrix is transferred from Riemannian to Euclidean space by tangent space mapping with the LogEig operation. Then, the learned HPD matrix is converted into a complex-valued vector, and a CV-3DCNN module is followed to learn contextual information to enhance feature representation. Finally, a softmax classifier is utilized to obtain the final result.

\begin{figure*}[ht]
	\centering
	\setlength{\fboxrule}{0.2pt}
	\setlength{\fboxsep}{0.01mm}
	\includegraphics[height=0.3\textheight]{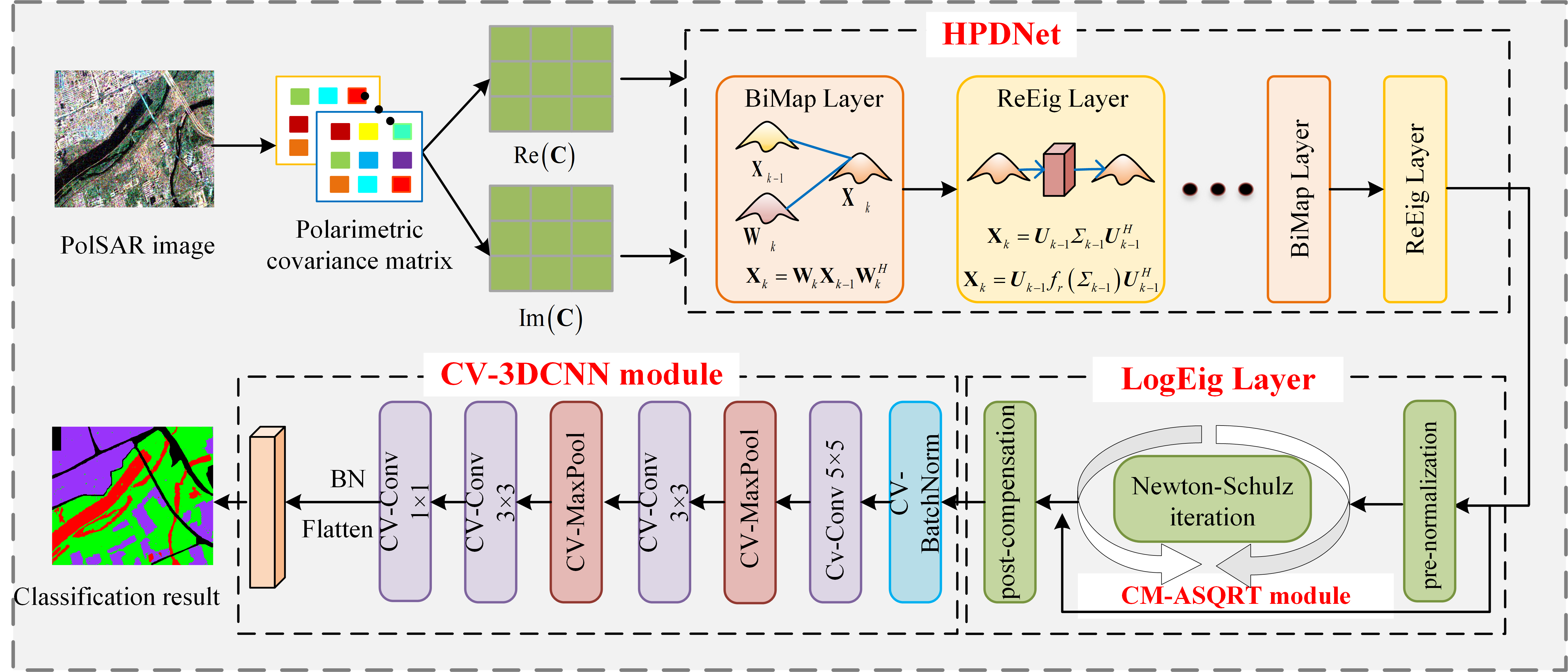}
	\caption{The framework of the proposed Riemannian complex HPD convolution network.}
	\label{frame}
\end{figure*}

\subsection{ PolSAR HPD matrix representation}
PolSAR system is imaging by emitting and receiving electromagnetic waves with four polarimetric modes. So, the scattering matrix is represented by
\begin{equation}
{\bf{S}} = \left[ {\begin{array}{*{20}{c}}
   {{S_{hh}}} & {{S_{hv}}}  \\
   {{S_{vh}}} & {{S_{vv}}}  \\
\end{array}} \right]
\end{equation}
where $S_{hh}$ is the echo of the antenna scattering in horizontal emitting and horizontal receiving mode. Under the assumption of reciprocity, $S_{hv}=S_{vh}$. Generally, $\textbf{S}$ can be vectorized as $k = \left[ {{S_{hh}},\sqrt 2 {S_{hv}},{S_{vv}}} \right]$. After multi-look processing, a covariance matrix can be achieved by
\begin{equation}
{\bf{C}} = \left[ {\begin{array}{*{20}{c}}
{{C_{11}}}&{{C_{12}}}&{{C_{13}}}\\
{{C_{21}}}&{{C_{22}}}&{{C_{23}}}\\
{{C_{31}}}&{{C_{32}}}&{{C_{33}}}
\end{array}} \right]
\end{equation}

Since each non-diagonal element is complex-valued data, the covariance matrix $\textbf{C}$ can be unfolded as:
\begin{equation}
 \tiny
{\bf{C}} = \left[ {\begin{array}{*{20}{c}}
{{C_{11}}}&{\Re \left( {{C_{12}}} \right)}&{\Re \left( {{C_{13}}} \right)}\\
{\Re \left( {{C_{21}}} \right)}&{{C_{22}}}&{\Re \left( {{C_{23}}} \right)}\\
{\Re \left( {{C_{31}}} \right)}&{\Re \left( {{C_{32}}} \right)}&{{C_{33}}}
\end{array}} \right] + j\cdot\left[ {\begin{array}{*{20}{c}}
0&{\Im \left( {{C_{12}}} \right)}&{\Im \left( {{C_{13}}} \right)}\\
{\Im \left( {{C_{21}}} \right)}&0&{\Im \left( {{C_{23}}} \right)}\\
{\Im \left( {{C_{31}}} \right)}&{\Im \left( {{C_{32}}} \right)}&0
\end{array}} \right]
\end{equation}

\subsection{ Complex HPD network}
Traditional SPD network in Riemannian space only learn the covariance matrix as a real matrix, which did not fully consider complex-matrix structure and characteristics of real part and imaginary part. To learn manifold structure of HPD matrix well, we design a complex HPD unfolding network to better learn the real and imaginary information of HPD matrix. The HPD unfolding network consists of complex HPD mapping layer, complex HPD Rectifying layer and LogEig layer. In addition, a fast eigenvalue decomposition method is designed for HPD matrices.

\emph{1) Complex HPD mapping layer}

\begin{figure}[ht]
	\centering
	\setlength{\fboxrule}{0.2pt}
	\setlength{\fboxsep}{0.01mm}
	\includegraphics[height=0.18\textheight]{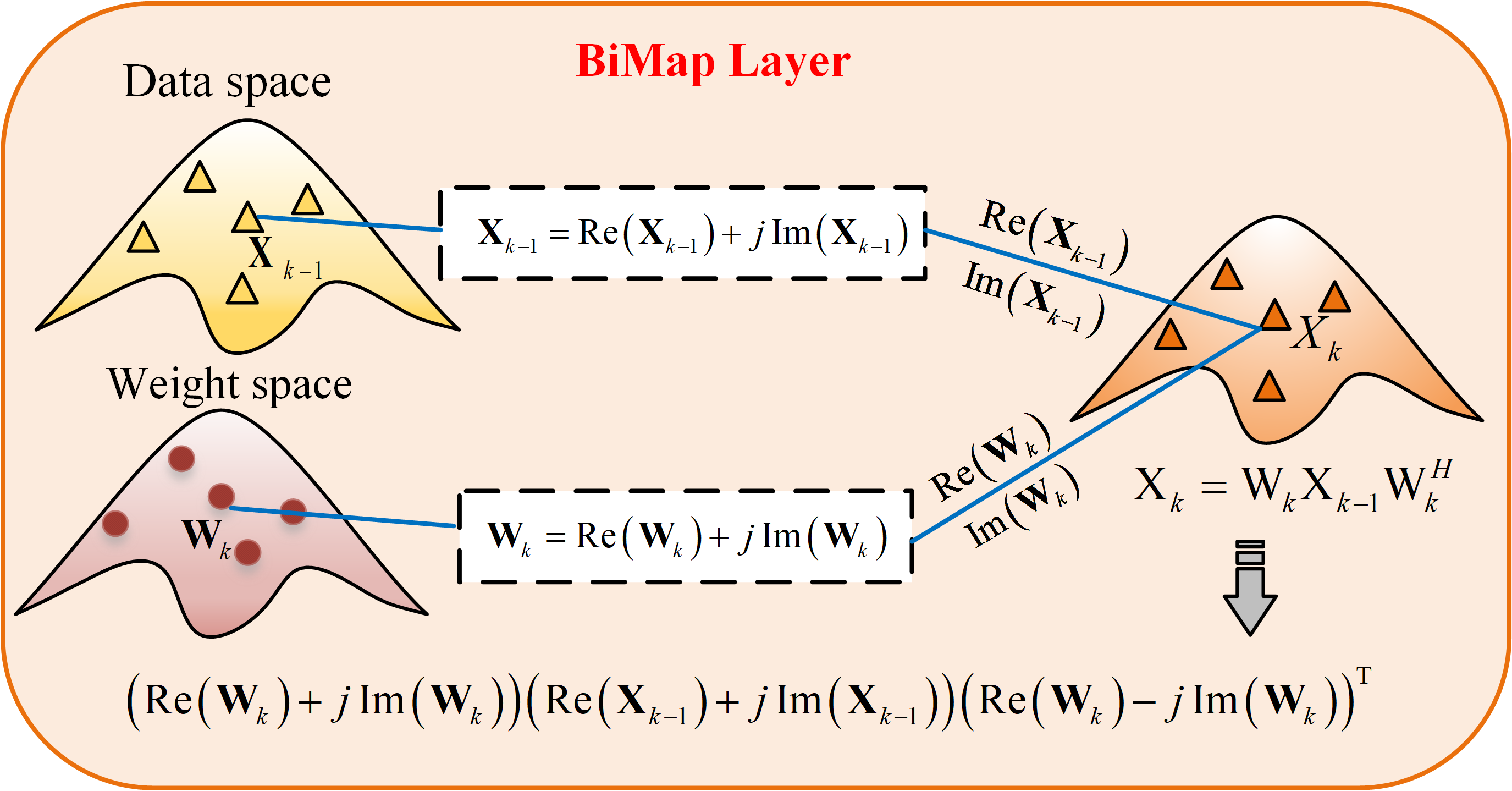}
	\caption{The complex HPD mapping layer.}
	\label{bi}
\end{figure}

The matrix mapping layer can map an HPD matrix from one HPD manifold to another. Here, considering that each covariance matrix is HPD, we unfold an HPD matrix in the addition of real and complex matrices. Given a complex HPD convolution kernel $W_k$, the complex HPD matrix mapping layer can be defined as

\begin{equation}\label{4}
\tiny
 \begin{array}{l}
{X_k} = {f_m}\left( {{W_k},{X_{k - 1}}} \right) = {W_k}{X_{k - 1}}W_k^H\\
 = \left( {\Re \left( {{W_k}} \right) + j\Im \left( {{W_k}} \right)} \right)\left( {\Re \left( {{X_{k - 1}}} \right) + j\Im \left( {{X_{k - 1}}} \right)} \right){\left( {\Re \left( {{W_k}} \right) - j\Im \left( {{W_k}} \right)} \right)^T}\\
 = \left( {\Re \left( {{W_k}} \right)\Re \left( {{X_{k - 1}}} \right)\Re {{\left( {{W_k}} \right)}^T} - \Im \left( {{W_k}} \right)\Im \left( {{X_{k - 1}}} \right)\Re {{\left( {{W_k}} \right)}^T}} \right.\\
\left. { + \Re \left( {{W_k}} \right)\Im \left( {{X_{k - 1}}} \right)\Im {{\left( {{W_k}} \right)}^T} + \Im \left( {{W_k}} \right)\Re \left( {{X_{k - 1}}} \right)\Im {{\left( {{W_k}} \right)}^T}} \right)\\
 + j\left( { - \Re \left( {{W_k}} \right)\Re \left( {{X_{k - 1}}} \right)\Im {{\left( {{W_k}} \right)}^T} + \Im \left( {{W_k}} \right)\Im \left( {{X_{k - 1}}} \right)\Im {{\left( {{W_k}} \right)}^T}} \right.\\
\left. { + \Re \left( {{W_k}} \right)\Im \left( {{X_{k - 1}}} \right)\Re {{\left( {{W_k}} \right)}^T} + \Im \left( {{W_k}} \right)\Re \left( {{X_{k - 1}}} \right)\Re {{\left( {{W_k}} \right)}^T}} \right)
\end{array}
\end{equation}
where $f_m$ is the mapping function, $\textbf{W}_k$ is a complex convolution kernel. $\textbf{W}_k$ can be unfolded as ${{\bf{W}}_k} = \Re \left( {{{\bf{W}}_k}} \right) + j \cdot \Im \left( {{{\bf{W}}_k}} \right)$. After unfolding mapping, the obtained output $\textbf{W}_k$ should also an HPD matrix\cite{wang2021symnet}. The HPD-based BiMap Layer can be shown in Fig.\ref{bi}.

\emph{2)Complex HPD Rectifying layer}

After matrix mapping layer, the original HPD matrix is converted to a new manifold feature space. However, the HPD mapping layer is similar to the linear mapping in SPDnet. A non-linear rectifying is necessary to enhance the discriminating ability of the mapping features. With non-linear rectifying, the original HPD matrix can be mapped to a new space with better separability. Here, we define a nonlinear function $f_r$ to rectify the result from the mapping layer, denoted by
\begin{equation}\label{e8}
\tiny
\begin{array}{l}
{X_k} = {f_r}\left( {{X_{k - 1}}} \right) = {U_{k - 1}}{f_r}\left( {{\Lambda _{k - 1}}} \right)U_{k - 1}^H\\
 = \left( {\Re \left( {{U_{k - 1}}} \right) + j\Im \left( {{U_{k - 1}}} \right)} \right){f_r}\left( {{\Lambda _{k - 1}}} \right){\left( {\Re \left( {{U_{k - 1}}} \right) - j\Im \left( {{U_{k - 1}}} \right)} \right)^T}\\
 = \left( {\Re \left( {{U_{k - 1}}} \right){f_r}\left( {{\Lambda _{k - 1}}} \right)\Re {{\left( {{U_{k - 1}}} \right)}^T} - \Im \left( {{U_{k - 1}}} \right){f_r}\left( {{\Lambda _{k - 1}}} \right)\Im {{\left( {{U_{k - 1}}} \right)}^T}} \right)\\
 + j\left( {\Im \left( {{U_{k - 1}}} \right){f_r}\left( {{\Lambda _{k - 1}}} \right)\Re {{\left( {{U_{k - 1}}} \right)}^T} + \Re \left( {{U_{k - 1}}} \right){f_r}\left( {{\Lambda _{k - 1}}} \right)\Im {{\left( {{U_{k - 1}}} \right)}^T}} \right)
\end{array}
\end{equation}
$f_r$ is defined based on the eigenvalue decomposition, where eigenvalues are rectified if they are less than a threshold. So, ${f_r}\left( {{\Lambda _{k - 1}}} \right) = {U_{k - 1}}\max \left( {\tau I,{\Lambda _{k - 1}}} \right)U_{k - 1}^H$. ${\Lambda _{k - 1}}$ is the eigenvalues and $\tau $ is the threshold. Then, $\max \left( {\tau I,{\Lambda _{k - 1}}} \right)$ can be written as:
\begin{equation}\label{e9}
\max {\left( {\tau I,{\Lambda _{k - 1}}} \right)_{ii}} = \left\{ {\begin{array}{*{20}{c}}
{{{\left( {{\Lambda _{k - 1}}} \right)}_{ii}},   if{\rm{  }}{{\left( {{\Lambda _{k - 1}}} \right)}_{ii}} > \tau }\\
{{{\left( {{\Lambda _{k - 1}}} \right)}_{ii}},   otherwise.       }
\end{array}} \right.
\end{equation}

\begin{figure}[ht]
	\centering
	\setlength{\fboxrule}{0.2pt}
	\setlength{\fboxsep}{0.01mm}
	\includegraphics[height=0.18\textheight]{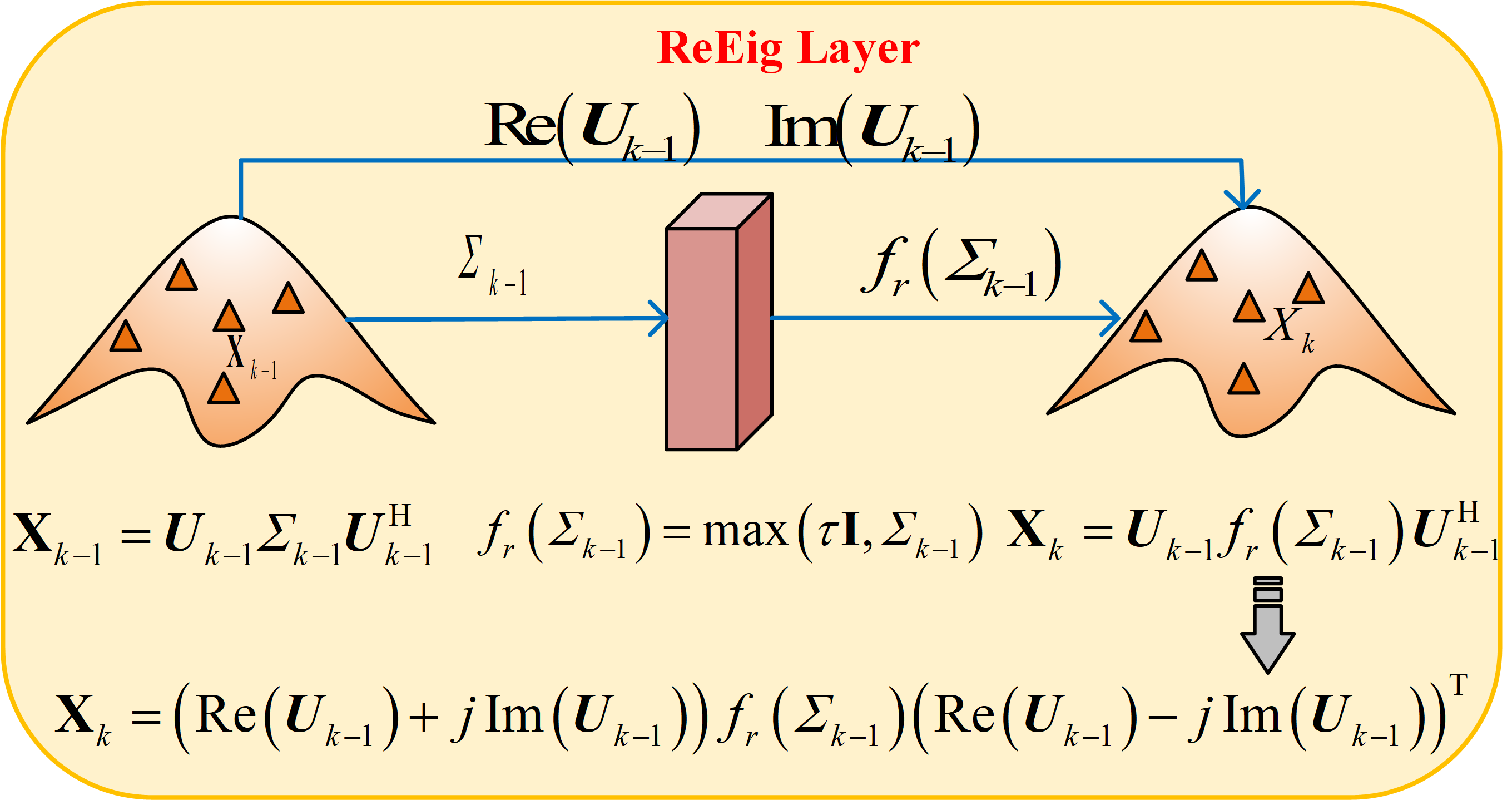}
	\caption{The complex HPD Rectifying layer.}
	\label{re}
\end{figure}

Here, we give $\tau$ as a positive number and above the smallest eigenvalue, which ensures the rectifying layer is the nonlinear operation. After rectifying, the resulting matrices are still HPD. The HPD-based ReEig Layer can be shown in Fig.\ref{re}.

The complex HPD mapping layer and rectifying layer can be considered as the convolution and ReLu layer similar to CNN. However, they transfer complex HPD matrix from one manifold to another. After applying multiple mapping and rectifying layers, a discriminating feature in Riemannian space can be achieved, which maintains the geometric structure of the PolSAR data and increases the discriminating ability of different classes.

\emph{3)Complex HPD LogEig layer}

After multiple layers of mapping and non-linear operations, the manifold HPD features are generated. Then, a fully connected operation is used to integrate all the features for classification. The existing classifier, such as the softmax classifier, is utilized in Euclidean space. To flatten the HPD matrices from Riemannian to Euclidean space, a logarithmic operation is defined to convert the HPD matrix from manifold space into tangent space, in which Euclidean operations can be utilized. Here, we define a complex HPD matrix logarithm function $f_{log}$ to convert manifold data to the flat tangent space, denoted by:
\begin{equation}\label{e10}
\begin{array}{l}
{X_k} = {f_{\log }}\left( {{X_{k - 1}}} \right) = \log m\left( {{X_{k - 1}}} \right)\\
{\kern 1pt} {\kern 1pt} {\kern 1pt} {\kern 1pt} {\kern 1pt} {\kern 1pt} {\kern 1pt} {\kern 1pt} {\kern 1pt} {\kern 1pt} {\kern 1pt} {\kern 1pt} {\kern 1pt} {\kern 1pt} {\kern 1pt} {\kern 1pt}  = {U_{k - 1}}diag\left( {\log\left( {{\Lambda _{k - 1}}} \right)} \right)U_{k - 1}^H
\end{array}
\end{equation}
where ${X_{k - 1}} = {U_{k - 1}}{\Lambda _{k - 1}}U_{k - 1}^H$ is the matrix's eigenvalue decomposition. $diag\left( {\log \left( {{\Lambda_{k - 1}}} \right)} \right)$ is to convert ${\log\left( {{\Lambda_{k - 1}}} \right)}$ to a diagonal matrix.

\emph{4)Fast eigenvalue decomposition}

Generally, SVD decomposition is utilized to obtain the eigenvalues and eigenvector. It can be observed that both the complex HPD Rectifying and LogEig layers need the SVD decomposition of the complex HPD matrix. However, for a complex HPD matrix, SVD decomposition is time-consuming since the complex operations of matrix inversion and trace are difficult to compute during the operation process. To address this issue, we utilize an ASQRT method\cite{li2018towards}, which is based on the Newton-Schulz iteration. The ASQRT method can greatly reduce computation time, since it only needs matrix multiplication instead of matrix inversion. These multiplication operations can be effective and fast to be conducted on GPU with parallel implementation. The SVD decomposition of the covariance matrix can be approximately calculated by Newton-Schulz iteration\cite{li2018towards}. To be specific, if we want to compute the square root $X$ of $C$. We can initialize ${X_0}=C$, and ${Z_0}=I$. Then, the coupled iteration can be calculated by :
\begin{equation}\label{e11}
\begin{array}{l}
{X_k} = \frac{1}{2}{X_{k - 1}}\left( {3I - {Z_{k - 1}}{X_{k - 1}}} \right)\\
{Z_k} = \frac{1}{2}\left( {3I - {Z_{k - 1}}{Y_{k - 1}}} \right){Z_{k - 1}}
\end{array}
\end{equation}

This procedure can obtain the approximation solution with a small number of iterations. However, this method is suitable for the real matrix. To expand them into the complex HPD matrix, we proposed a complex matrix-based ASQRT method(CM-ASQRT), which unfolds these equations by defining each matrix as an HPD matrix with real and imaginary parts. That is, assume ${\bf{X_{k-1}}} = {\bf{C}} = \Re \left( {\bf{C}} \right) + j\Im \left( {\bf{C}} \right)$, and ${\bf{Z}} = \Re \left( {\bf{I}} \right) + j{\bf{0}}$. By replacing them into Eq.(\ref{e11}), the real and imaginary parts of ${{\bf{X}}_k}$ can be obtained as shown in Eq.(\ref{e1}) and Eq.(\ref{e2}), respectively.

Similarly, the real and imaginary part of ${{\bf{Z}}_k}$ can be derived by Eq.(\ref{e3}) and Eq.(\ref{e4}), respectively.

After several iterations, the SVD decomposition can be approximately obtained. It has been demonstrated\cite{li2018towards} that no more than 5 iterations can obtain good performance under a deep learning architecture. Here, we select 5 as the iteration number.

\begin{equation}\label{e1}
\tiny
\begin{array}{l}
 \Re \left( {{{\bf{X}}_k}} \right) = \frac{3}{2}\Re \left( {{{\bf{X}}_{k - 1}}} \right)\Re \left( {\bf{I}} \right) - \frac{1}{2}\Re \left( {{{\bf{X}}_{k - 1}}} \right)\Re \left( {{{\bf{Z}}_{k - 1}}} \right)\Re \left( {{{\bf{X}}_{k - 1}}} \right) \\
 {\rm{              }} + \frac{1}{2}\Re \left( {{{\bf{X}}_{k - 1}}} \right)\Im \left( {{{\bf{Z}}_{k - 1}}} \right)\Im \left( {{{\bf{X}}_{k - 1}}} \right) + \frac{1}{2}\Im \left( {{{\bf{X}}_{k - 1}}} \right)\Re \left( {{{\bf{Z}}_{k - 1}}} \right)\Im \left( {{{\bf{X}}_{k - 1}}} \right) \\
 {\rm{              }} + \frac{1}{2}\Im \left( {{{\bf{X}}_{k - 1}}} \right)\Im \left( {{{\bf{Z}}_{k - 1}}} \right)\Re \left( {{{\bf{X}}_{k - 1}}} \right) \\
 \end{array}
 \end{equation}

\begin{equation}   \label{e2}
\tiny
\begin{array}{l}
 \Im \left( {{{\bf{X}}_k}} \right) = 3\Im \left( {{{\bf{X}}_{k - 1}}} \right)\Re \left( {\bf{I}} \right) - \frac{1}{2}\Re \left( {{{\bf{X}}_{k - 1}}} \right)\Re \left( {{{\bf{Z}}_{k - 1}}} \right)\Im \left( {{{\bf{X}}_{k - 1}}} \right) \\
 {\rm{              }} - \frac{1}{2}\Re \left( {{{\bf{X}}_{k - 1}}} \right)\Im \left( {{{\bf{Z}}_{k - 1}}} \right)\Re \left( {{{\bf{X}}_{k - 1}}} \right) - \frac{1}{2}\Im \left( {{{\bf{X}}_{k - 1}}} \right)\Re \left( {{{\bf{Z}}_{k - 1}}} \right)\Re \left( {{{\bf{X}}_{k - 1}}} \right) \\
 {\rm{             }} + \frac{1}{2}\Im \left( {{{\bf{X}}_{k - 1}}} \right)\Im \left( {{{\bf{Z}}_{k - 1}}} \right)\Im \left( {{{\bf{X}}_{k - 1}}} \right) \\
 \end{array}
\end{equation}

\begin{equation}\label{e3}
\tiny
\begin{array}{l}
 \Re \left( {{{\bf{Z}}_k}} \right) = \frac{3}{2}\Re \left( {\bf{I}} \right)\Re \left( {{{\bf{Z}}_{k - 1}}} \right) - \frac{1}{2}\Re \left( {{{\bf{Z}}_{k - 1}}} \right)\Re \left( {{{\bf{X}}_{k - 1}}} \right)\Re \left( {{{\bf{Z}}_{k - 1}}} \right) \\
 {\kern 1pt} {\kern 1pt} {\kern 1pt} {\kern 1pt} {\kern 1pt} {\kern 1pt} {\kern 1pt} {\kern 1pt} {\kern 1pt} {\kern 1pt} {\kern 1pt} {\kern 1pt} {\kern 1pt} {\kern 1pt} {\kern 1pt} {\kern 1pt} {\kern 1pt} {\kern 1pt} {\kern 1pt} {\kern 1pt} {\kern 1pt} {\kern 1pt} {\kern 1pt} {\kern 1pt} {\kern 1pt} {\kern 1pt} {\kern 1pt} {\kern 1pt} {\kern 1pt} {\kern 1pt} {\kern 1pt} {\kern 1pt} {\kern 1pt} {\kern 1pt} {\kern 1pt} {\kern 1pt} {\kern 1pt} {\kern 1pt} {\kern 1pt} {\kern 1pt} {\kern 1pt}  + \frac{1}{2}\Re \left( {{{\bf{Z}}_{k - 1}}} \right)\Im \left( {{{\bf{X}}_{k - 1}}} \right)\Im \left( {{{\bf{X}}_{k - 1}}} \right) + \frac{1}{2}\Im \left( {{{\bf{Z}}_{k - 1}}} \right)\Re \left( {{{\bf{X}}_{k - 1}}} \right)\Im \left( {{{\bf{Z}}_{k - 1}}} \right) \\
 {\kern 1pt} {\kern 1pt} {\kern 1pt} {\kern 1pt} {\kern 1pt} {\kern 1pt} {\kern 1pt} {\kern 1pt} {\kern 1pt} {\kern 1pt} {\kern 1pt} {\kern 1pt} {\kern 1pt} {\kern 1pt} {\kern 1pt} {\kern 1pt} {\kern 1pt} {\kern 1pt} {\kern 1pt} {\kern 1pt} {\kern 1pt} {\kern 1pt} {\kern 1pt} {\kern 1pt} {\kern 1pt} {\kern 1pt} {\kern 1pt} {\kern 1pt} {\kern 1pt} {\kern 1pt} {\kern 1pt} {\kern 1pt} {\kern 1pt} {\kern 1pt} {\kern 1pt} {\kern 1pt} {\kern 1pt} {\kern 1pt} {\kern 1pt} {\kern 1pt}  + \frac{1}{2}\Im \left( {{{\bf{Z}}_{k - 1}}} \right)\Im \left( {{{\bf{X}}_{k - 1}}} \right)\Re \left( {{{\bf{Z}}_{k - 1}}} \right) \\
 \end{array}
 \end{equation}

 \begin{equation}\label{e4}
 \tiny
\begin{array}{l}
 \Im \left( {{{\bf{Z}}_k}} \right) = \frac{3}{2}\Re \left( {\bf{I}} \right)\Im \left( {{{\bf{Z}}_{k - 1}}} \right) - \frac{1}{2}\Re \left( {{{\bf{Z}}_{k - 1}}} \right)\Re \left( {{{\bf{X}}_{k - 1}}} \right)\Im \left( {{{\bf{Z}}_{k - 1}}} \right) \\
 {\kern 1pt} {\kern 1pt} {\kern 1pt} {\kern 1pt} {\kern 1pt} {\kern 1pt} {\kern 1pt} {\kern 1pt} {\kern 1pt} {\kern 1pt} {\kern 1pt} {\kern 1pt} {\kern 1pt} {\kern 1pt} {\kern 1pt} {\kern 1pt} {\kern 1pt} {\kern 1pt} {\kern 1pt} {\kern 1pt} {\kern 1pt} {\kern 1pt} {\kern 1pt} {\kern 1pt} {\kern 1pt} {\kern 1pt} {\kern 1pt} {\kern 1pt} {\kern 1pt} {\kern 1pt} {\kern 1pt} {\kern 1pt} {\kern 1pt} {\kern 1pt} {\kern 1pt} {\kern 1pt} {\kern 1pt} {\kern 1pt} {\kern 1pt} {\kern 1pt}  - \frac{1}{2}\Re \left( {{{\bf{Z}}_{k - 1}}} \right)\Im \left( {{{\bf{X}}_{k - 1}}} \right)\Re \left( {{{\bf{Z}}_{k - 1}}} \right) - \frac{1}{2}\Im \left( {{{\bf{Z}}_{k - 1}}} \right)\Re \left( {{{\bf{X}}_{k - 1}}} \right)\Re \left( {{{\bf{Z}}_{k - 1}}} \right) \\
 {\kern 1pt} {\kern 1pt} {\kern 1pt} {\kern 1pt} {\kern 1pt} {\kern 1pt} {\kern 1pt} {\kern 1pt} {\kern 1pt} {\kern 1pt} {\kern 1pt} {\kern 1pt} {\kern 1pt} {\kern 1pt} {\kern 1pt} {\kern 1pt} {\kern 1pt} {\kern 1pt} {\kern 1pt} {\kern 1pt} {\kern 1pt} {\kern 1pt} {\kern 1pt} {\kern 1pt} {\kern 1pt} {\kern 1pt} {\kern 1pt} {\kern 1pt} {\kern 1pt} {\kern 1pt} {\kern 1pt} {\kern 1pt} {\kern 1pt} {\kern 1pt} {\kern 1pt} {\kern 1pt} {\kern 1pt} {\kern 1pt} {\kern 1pt}  + \frac{1}{2}\Im \left( {{{\bf{Z}}_{k - 1}}} \right)\Im \left( {{{\bf{X}}_{k - 1}}} \right)\Im \left( {{{\bf{Z}}_{k - 1}}} \right) \\
 \end{array}
\end{equation}


\subsection{Remianian-to-Euclidean complex-valued 3D-CNN enhanced classification}

After log-Eig operation, the complex HPD matrix in Riemannian space can be projected to the tangent space, in which Euclidean operations can be utilized directly. Thus, the HPD covariance matrix can be learned from Riemannian to Euclidean space. Then, a CV-3DCNN network architecture is utilized to learn contextual information about complex-valued data in Euclidean space.
In this model, the HPD matrix in each pixel is converted into a complex-valued vector. An image block can be represented by a complex-valued 3D tensor, noted by $I \in {\mathcal{C}^{N \times N \times 6}}$. $N\times N$ is the image block size and 6 is the scattering channel number of complex matrix. Then, CV-3DCNN can learn both spatial and scattering information simultaneously. It consists of CV-3D convolution, activation and pooling layers.

1)\textit{CV-3D convolution layer}: For the input image block \textbf{I}, for $i$th layer convolution, assuming a set of filter bands is defined as $w \in \mathcal{C}$ and the bias is ${b_i} \in \mathcal{C}$, where the size of filter bands is ${M_i} \times {M_i} \times {R_i} \times {K_i}$. ${M_i} \times {M_i}$ is the kernel size in spatial dimension, and ${R_i}$ is the kernel size in channel dimension. ${K_i}$ is the number of filters. Then, the jth($j \in \left( {0, \ldots ,{K_i}} \right)$) feature map in $i$th layer can be calculated by:
\begin{equation}\label{e}
\begin{array}{l}
{f_{ij}} = \Re \left( {{w_{ij}}} \right) * \Re \left( {{f_{i - 1}}} \right) - \Im \left( {{w_{ij}}} \right) * \Im \left( {{f_{i - 1}}} \right)\\
{\kern 1pt} {\kern 1pt} {\kern 1pt} {\kern 1pt} {\kern 1pt} {\kern 1pt} {\kern 1pt} {\kern 1pt} {\kern 1pt} {\kern 1pt} {\kern 1pt} {\kern 1pt} {\kern 1pt} {\kern 1pt} {\kern 1pt} {\kern 1pt} {\kern 1pt} {\kern 1pt} {\kern 1pt} {\kern 1pt} {\kern 1pt} {\kern 1pt}  + j \cdot \left( {\Re \left( {{w_{ij}}} \right) * \Im \left( {{f_{i - 1}}} \right) - \Im \left( {{w_{ij}}} \right) * \Re \left( {{f_{i - 1}}} \right)} \right)\\
{\kern 1pt} {\kern 1pt} {\kern 1pt} {\kern 1pt} {\kern 1pt} {\kern 1pt} {\kern 1pt} {\kern 1pt} {\kern 1pt} {\kern 1pt} {\kern 1pt} {\kern 1pt} {\kern 1pt} {\kern 1pt} {\kern 1pt} {\kern 1pt} {\kern 1pt} {\kern 1pt} {\kern 1pt} {\kern 1pt} {\kern 1pt}  + \Re \left( {{b_{ij}}} \right) + j \cdot \left( {\Im \left( {{b_{ij}}} \right)} \right)
\end{array}
\end{equation}

According to CV-3D convolution, the later feature maps are connected to various polarimetric data in the previous layers, thus learning multiple scattering features.

2)\textit{Activation layer}: Each convolutional layer is the linear conversion of complex data, then a nonlinear activation function is needed to learn complex nonlinear transformation. The complex-valued Relu is defined as:
\begin{equation}\label{ex}
{\mathop{\rm Re}\nolimits} {\rm{LU}}\left( f \right) = {\mathop{\rm Re}\nolimits} {\rm{LU}}\left( {\Re \left( f \right)} \right) + j \cdot {\mathop{\rm Re}\nolimits} {\rm{LU}}\left( {\Im \left( f \right)} \right)
\end{equation}

3)\textit{Pooling layer}: Pooling layer can reduce the data dimension and fuse data features by down-sampling operation. Similarly, the real pooling operation can be expanded to Complex data domain, defined as:
\begin{equation}\label{ee4}
Pooling\left( f \right) = Pooling\left( {\Re \left( f \right)} \right) + j \cdot Pooling\left( {\Im \left( f \right)} \right)
\end{equation}

 4) \textit{Fully connection and softmax classification}

Fully connected layer attempts to flatten the CV features and convert the CV feature to real-valued features. Here, we flatten the CV cube as a CV vector. Then, real and imaginary parts are extracted respectively and connected together. Then, the real-valued feature maps are fed into a fully connected layer. Then a softmax classifier is applied and the MRF is utilized to smooth the classification result. The cross entropy loss is utilized to optimize the network. 

\subsection{Discussion}

\textbf{Differences between the proposed method and SPDnet:} There are four key points that distinguish our method from SPDnet. 1) the proposed complex HPD unfolding network is defined for PolSAR images for the first time, in which the PolSAR data itself is an HPD matrix endowed in Riemannian manifold, while the SPDnet is defined for natural images, which needs to convert features to a real SPD matrix. 2) We define an HDP unfolding network, which fully considers the real and imaginary matrices of the PolSAR covariance matrix, while SPDnet only considers a real matrix. 3) A fast eigenvalue decomposition method is designed for the HPD matrix in the ReEig and LogEig layers, named a revised complex HPD iteration model, which can accelerate the HPD network learning using parallel computing. 4) Our method combines the HPDnet in Riemannian space and complex-valued 3DCNN network in Euclidean space together, which can not only learn geometry features of PolSAR matrix in manifold space but also contextual semantic in Euclidean space. SPDnet only learns matrix transformation in the Riemannian space, while ignoring contextual relationship among SPD matrices.

\section{Experimental results and analysis}

\subsection{Experimental data and settings}
1)Experimental data

In this study, three PolSAR images were employed, each capturing real-world ground objects from distinct regions. These images were acquired from different satellites and frequency bands, covering the areas of Xi’an, Oberpfaffenhofen, and Flevoland2, respectively. Detailed information for each dataset is provided below.

1)Xi'an Dataset: The dataset consists of a C-band fully polarized image acquired over the Xi ’an region with the SIR-C/X-SAR system. The image has a spatial resolution of $8 \times 8$ meters and dimensions of $512 \times 512$ pixels. The captured area includes various types of land cover such as water, buildings, and grass. The Pauli RGB composite and the corresponding label map are presented in Figs.\ref{f10} (a) and (b), respectively.

2)Oberpfaffenhofen Dataset: This dataset was collected from the Oberpfaffenhofen region and featured full-polarimetric L-band SAR data acquired by the E-SAR system from the German Aerospace Center. The image has a spatial resolution of approximately $3 \times  3.2$ meters and dimensions of $1300 \times 1200$ pixels. The main types of land cover in this area include bare ground, forest, buildings, farmland, and roads. The Pauli RGB image and its corresponding label map are displayed in Figs. \ref{f11} (a) and (b), respectively.

3)Flevoland Dataset: The final dataset was collected by RADARSAT-2 and consists of C-band PolSAR data over Flevoland, Netherlands. This image consists of $1400 \times 1200$ pixels. The Pauli RGB image and its corresponding ground truth map are shown in Figs. \ref{f12} (a) and (b), respectively. This dataset includes four land cover classes: water, urban areas, woodland, and cropland.

\begin{figure}
	\centerline{\includegraphics[scale=0.3]{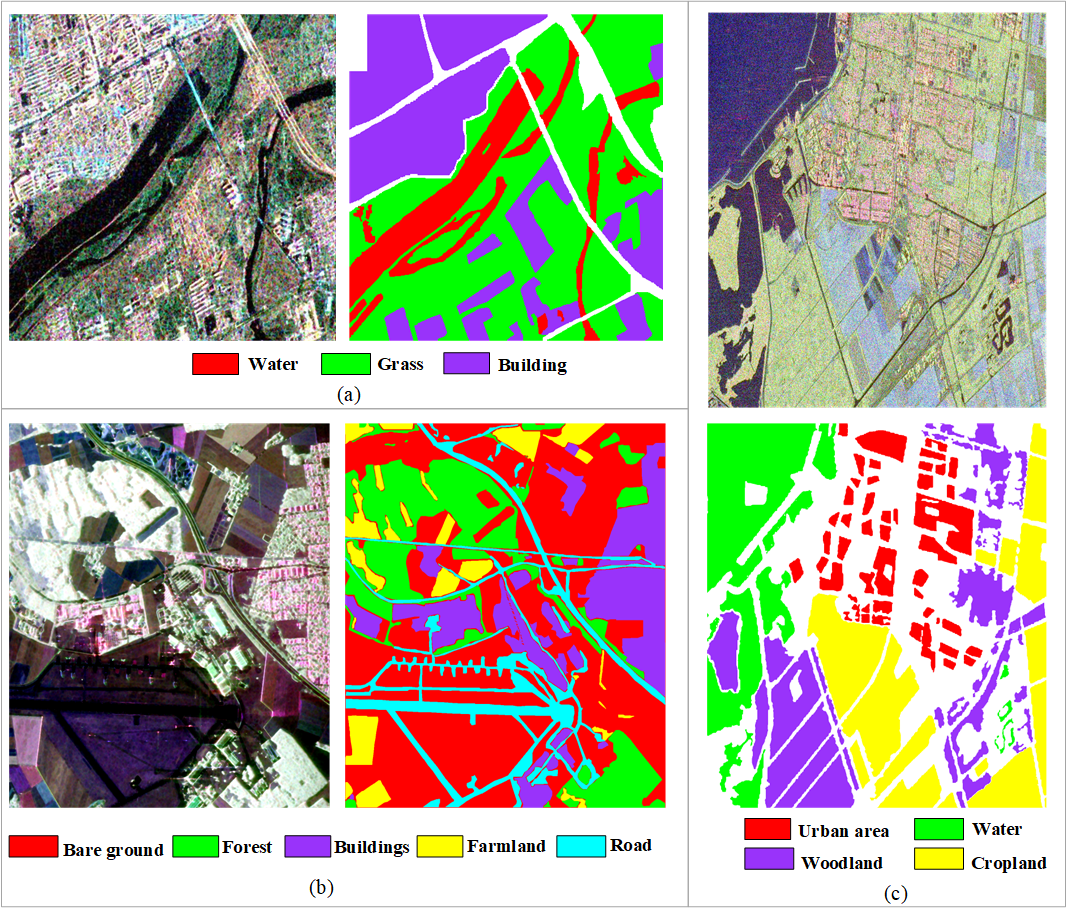}}
	\caption{Three datasets with their PauliRGB images and corresponding label maps. (a) Xi'an data set; (b) Oberpfaffenhofen data set; (c) Flevoland data set.}
	\label{fig-gt}
\end{figure}

2) Experimental settings

The experiments in this study were carried out using the PyTorch deep learning framework (version 1.6.0). All experiments were performed on a Windows 10 system equipped with an Intel Core i7-10700F processor, 64 GB of RAM, and an NVIDIA GeForce RTX 3070 GPU.

The parameters of the network model are configured as follows: a learning rate of 0.005 and a training process spanning 50 iterations. Training is performed using the Adam optimizer. The dataset is split into 10\% as the training set and 90\% for testing, with the training samples randomly selected. To assess the performance of the proposed method, several evaluation metrics are computed, including class precision, overall accuracy (OA), average accuracy (AA), the Kappa coefficient, and the confusion matrix.

To evaluate the effectiveness of the proposed method, comparisons are made against other leading PolSAR image classification techniques, including CV-CNN\cite{9323621}, Super\_RF\cite{RN36}, DFGCN\cite{9274334}, AMS-MESL\cite{10097620}, PolMPCNN\cite{9424197} and HybridCVnet\cite{RN109} methods. The CV-CNN method converts the complex matrix into a complex-valued vector for processing. The Super\_RF approach extracts statistical texture and polarization features to generate superpixels, and then the random forest model is applied for classification. The DFGCN method introduces a hybrid metric along with a fuzzy weighted graph convolutional network to capture contextual information for PolSAR image classification. Meanwhile, the AMS-MESL method constructs features as covariance matrices, applies an SPD manifold metric to learn geometric characteristics, and performs classification using Euclidean metrics. The PolMPCNN method employs a multi-channel network to learn scattering features by adaptively fusing multiple feature types. Finally, the HybridCVnet method combines CV-CNN and CV-ViT architectures to leverage complex-valued information for PolSAR classification.

\subsection{Experimental results of Xi'an data set}

The classification results of different methods on the Xi'an dataset are presented in Figs. 1(b)-(h), corresponding to CV-CNN, Super\_RF, DFGCN, AMS-MESL, PolMPCNN, HybridCVnet and the proposed HPD\_CNN, respectively. The ground truth is given in Fig.(a) for reference.
As shown in Figure 1(b), the CV-CNN method shows good performance in classifying the water category, while it still struggles with some noisy classes in the grass and building categories. The Super\_RF method (Figure 1(c)) also produces some misclassifications in urban and water regions due to block effect of superpixels. The DFGCN and AMS-MESL methods (Figs.1(e) and (f)) exhibit lower classification accuracy for the grass category compared to the previous methods, and the classification map contains noticeable noise. The PolMPCNN method in (g) can improve the region homogeneity and reduce speckle noise, while some edge pixels are confused. The HybridCVnet method can achieve good performance with complex-valued information, while some edge misclassificaitons still exist.
In contrast, the proposed method (Figure 1(h)) effectively learns the covariance matrix and high-level features, significantly enhancing classification accuracy. It can not only reduce noises but also improve regional consistency.

Table \ref{t1} compares the classification performance of different methods in the Xi'an dataset. The proposed method outperforms the benchmark algorithms, improving overall accuracy (OA) by 4.53\%, 10\%, 9\%, 6.78\%, 2.89\% and 2.34\%, respectively. Specifically, CV-CNN performs well in both water and building categories, achieving 94.55\% and 93.81\%, while the grass category remains low. The Super\_RF shows lowest accuracy in water class due to superpixels. The DFGCN shows poor performance in both water and building due to noisy result. The AMS-MESL method struggles with lower accuracy in all three categories. The PolMPCNN exhibits good performance in both water and building classes due to the ability to learn multiple channels, while the grass class is low due to misclassification. The HybridCVnet shows obvious improvements in three classes, while they are still lower than our method.
Compared to other classification algorithms, the proposed method obtains the highest overall accuracy (96.90\%) and Kappa coefficient (95.56\%).

 \begin{figure*}
	\centering
	\setlength{\fboxrule}{0.5pt}
	\setlength{\fboxsep}{0.01mm}
	\subfigure[]{\fbox{\includegraphics[height=0.14\textheight]{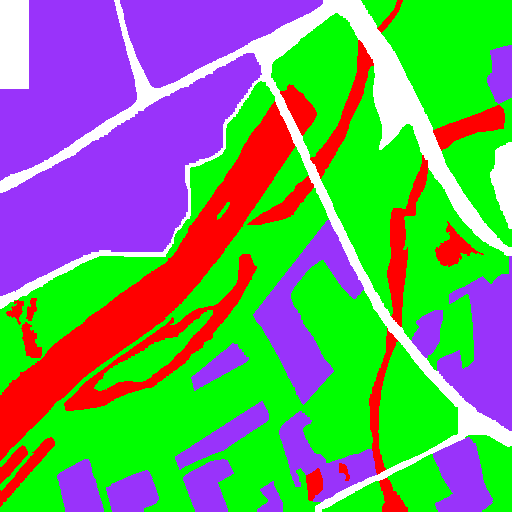}}}
	\subfigure[]{\fbox{\includegraphics[height=0.14\textheight]{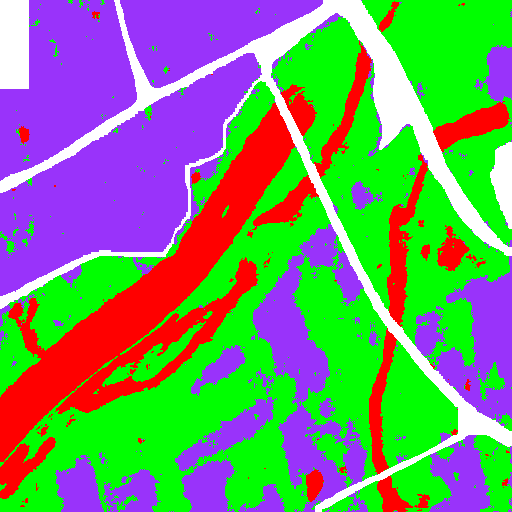}}}
	\subfigure[]{\fbox{\includegraphics[height=0.14\textheight]{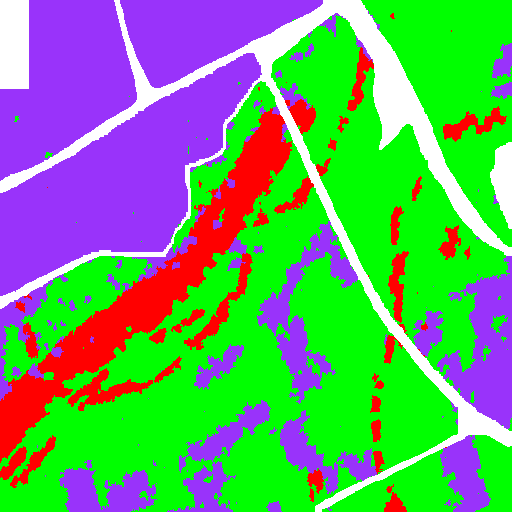}}}
	\subfigure[]{\fbox{\includegraphics[height=0.14\textheight]{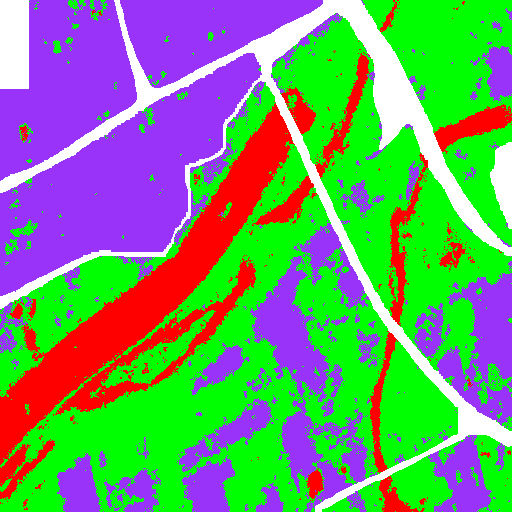}}}\\
	\subfigure[]{\fbox{\includegraphics[height=0.14\textheight]{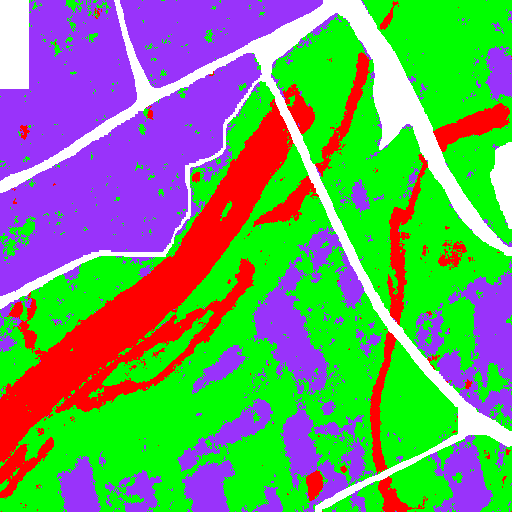}}}
	\subfigure[]{\fbox{\includegraphics[height=0.14\textheight]{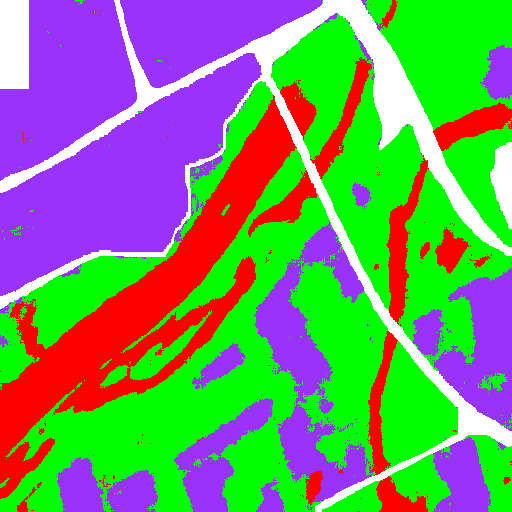}}}
     \subfigure[]{\fbox{\includegraphics[height=0.14\textheight]{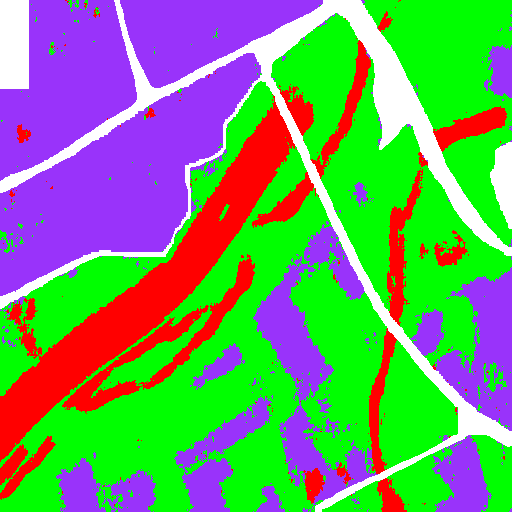}}}
    \subfigure[]{\fbox{\includegraphics[height=0.14\textheight]{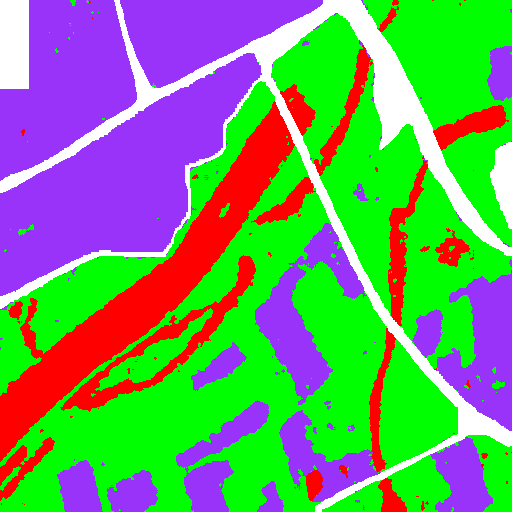}}}\\	
	\includegraphics[height=0.04\textheight]{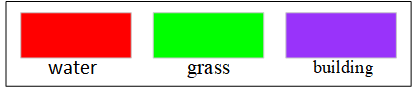}
	\caption{Classification results of different method on Xi'an area; (a)The label map; (b) CV-CNN;  (c) Super\_RF; (d) DFGCN; (e) AMS-MESL; (f) PolMPCNN; (g)HybridCVNet; (h)Proposed HPD\_CNN.}
	\label{f10}
\end{figure*}

\begin{table*}
\footnotesize
\begin{center}
\caption
{ \label{t1}
 Classification accuracy of different methods on Xi'an Data Set(\%).}
\begin{tabular}{lcccccccc}
\hline
class&CVCNN&Super\_RF&DFGCN&AMS-MESL&PolMPCNN&HybridCVnet&proposed\\
\hline
water&94.55&80.94&82.93&88.99&\textbf{95.52}&92.01&93.06\\
grass&90.68&91.23&90.89&90.35&90.95&94.99&\textbf{97.32}\\
building&93.81&83.34&85.79&90.27&97.68&95.05&\textbf{97.94}\\
OA&92.37&86.90&87.90&90.12&94.01&94.56&\textbf{96.90}\\
AA&93.01&85.17&86.54&89.87&94.71&94.02&\textbf{96.11}\\
Kappa&87.51&78.06&87.75&83.68&90.25&91.02&\textbf{95.56}\\
\hline
\end{tabular}
\end{center}
\end{table*}

\subsection{Experimental results of Oberpfaffenhofen data set}

\begin{figure*}
	\centering
	\setlength{\fboxrule}{0.5pt}
	\setlength{\fboxsep}{0.01mm}
\subfigure[]{\fbox{\includegraphics[height=0.16\textheight]{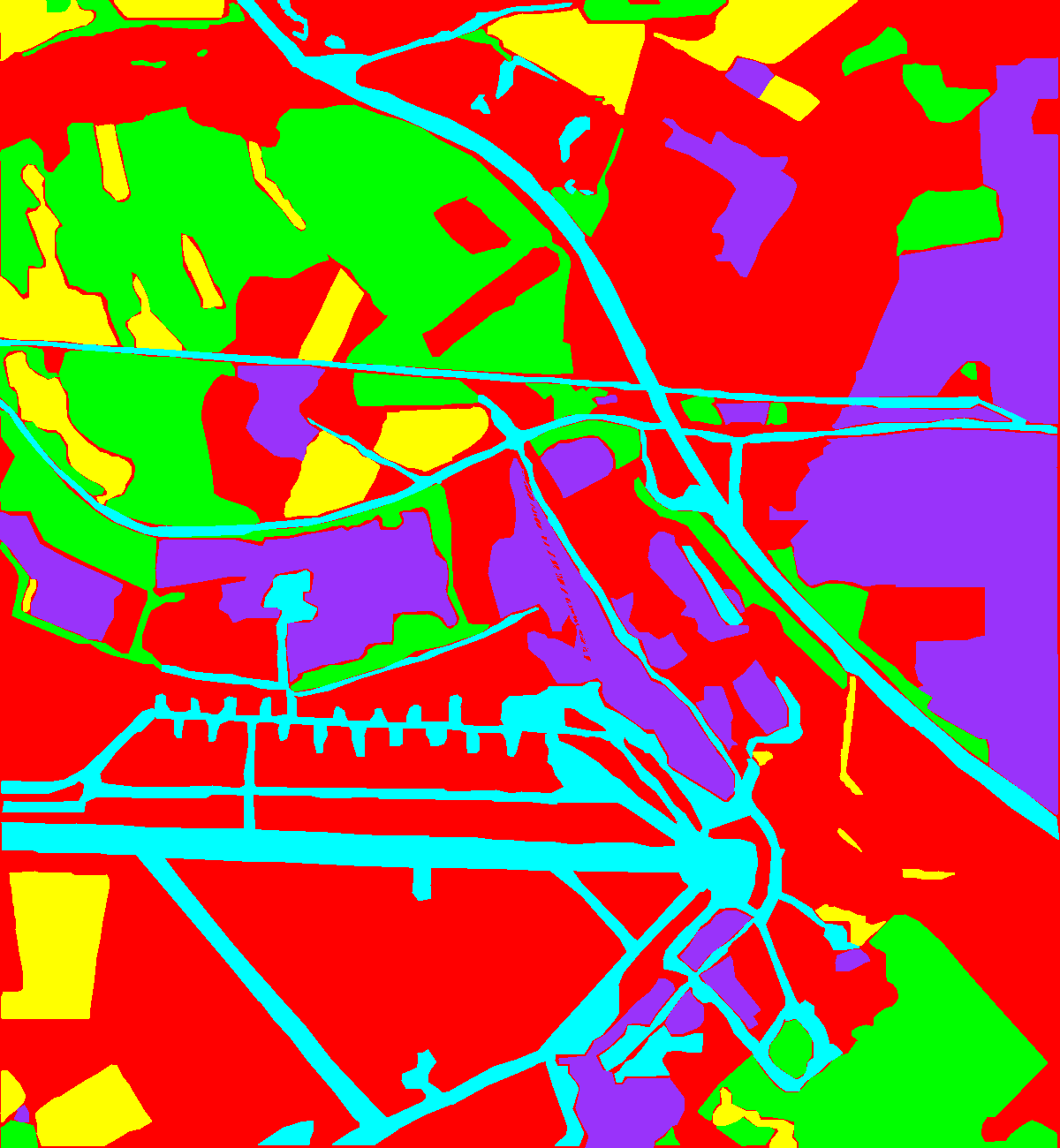}}}
	\subfigure[]{\fbox{\includegraphics[height=0.16\textheight]{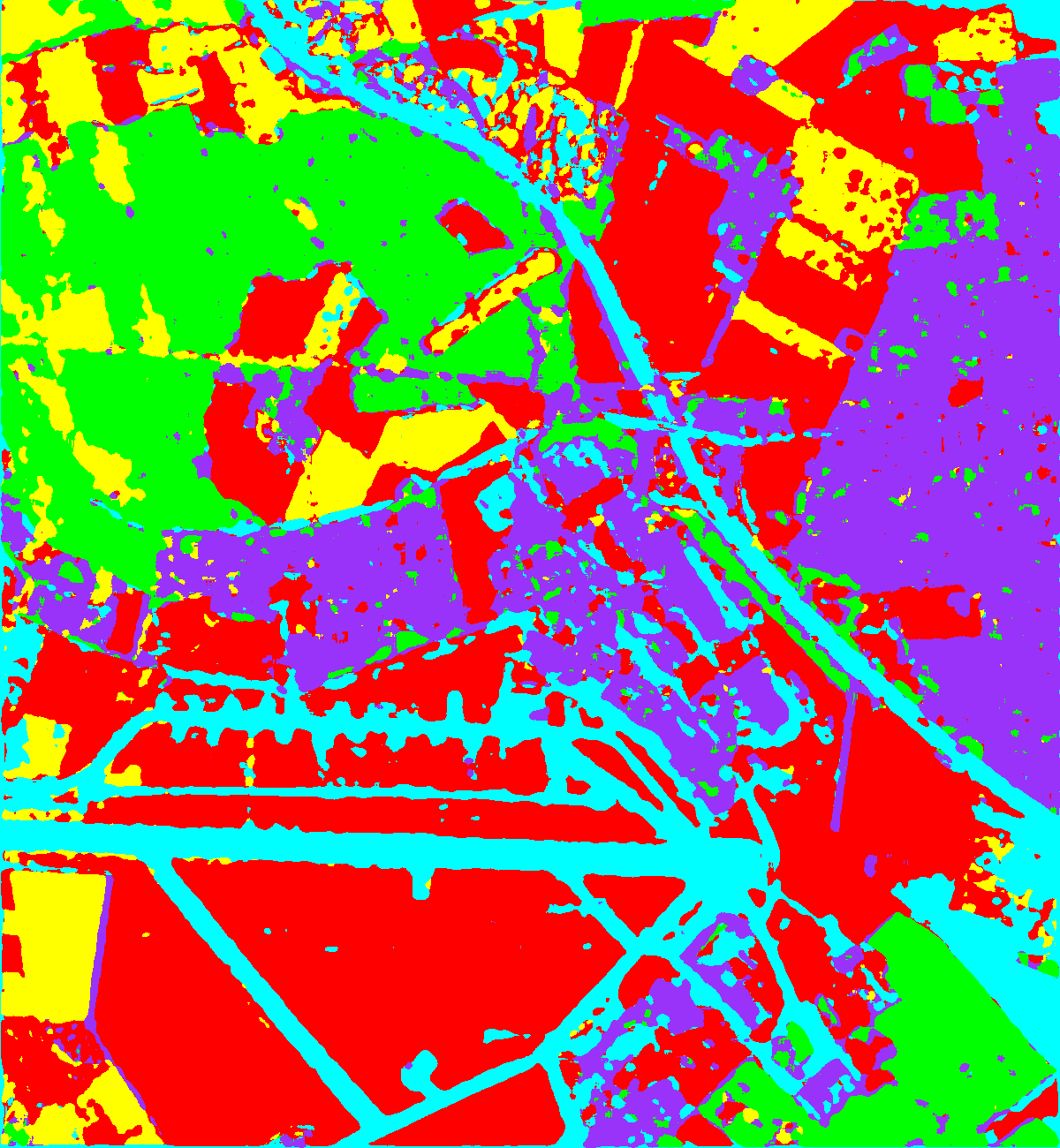}}}
	\subfigure[]{\fbox{\includegraphics[height=0.16\textheight]{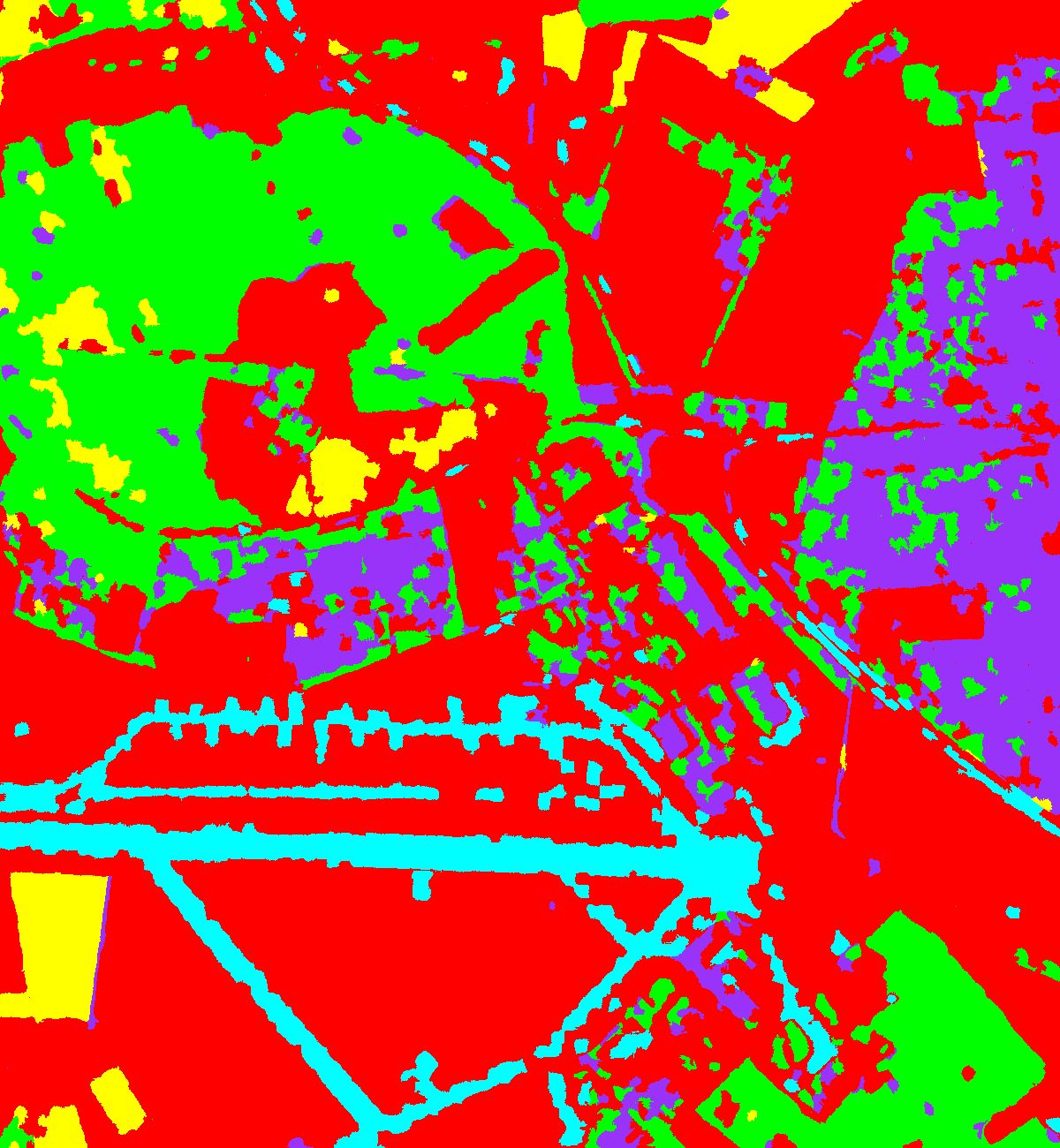}}}
	\subfigure[]{\fbox{\includegraphics[height=0.16\textheight]{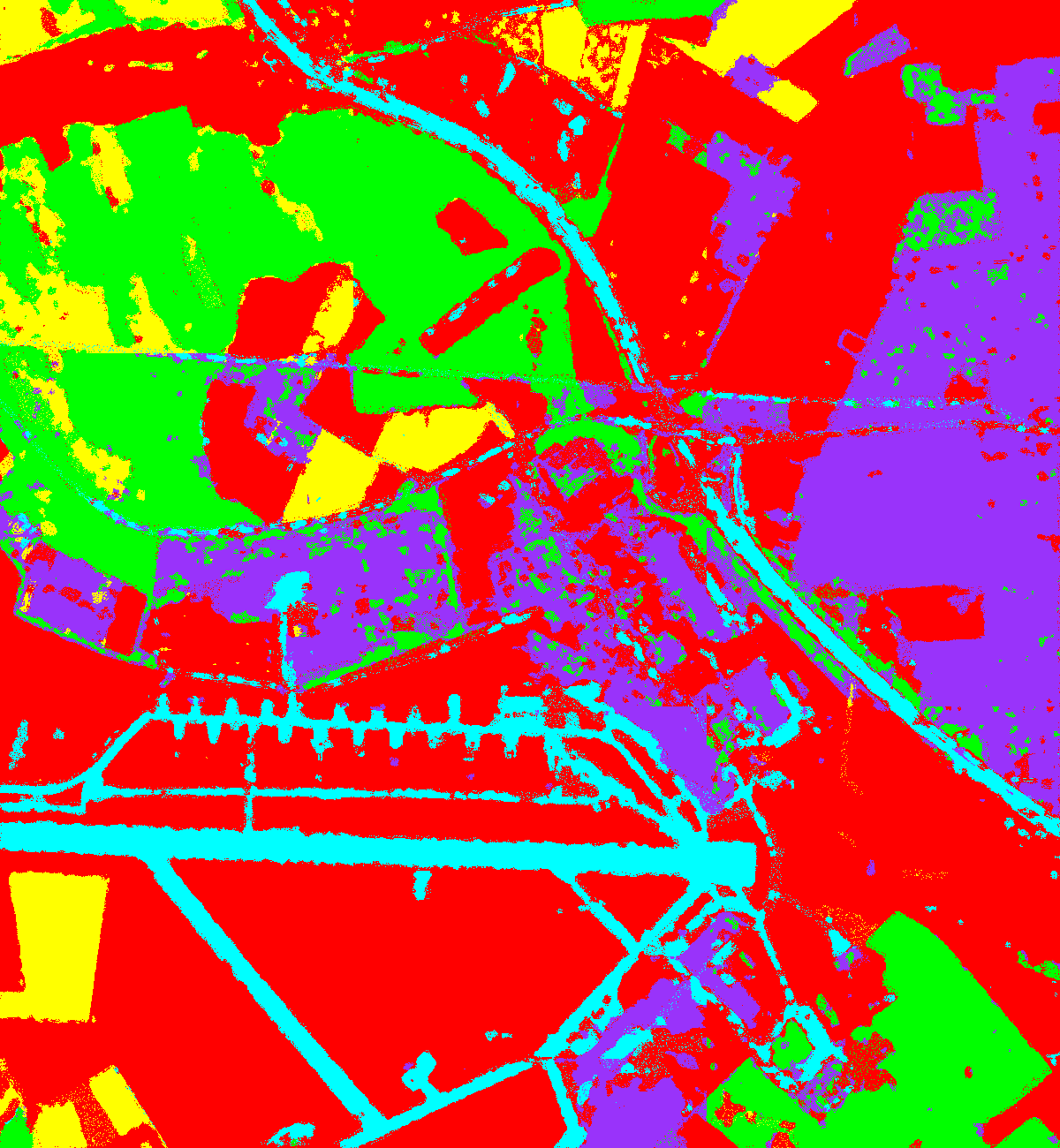}}}\\
	\subfigure[]{\fbox{\includegraphics[height=0.16\textheight]{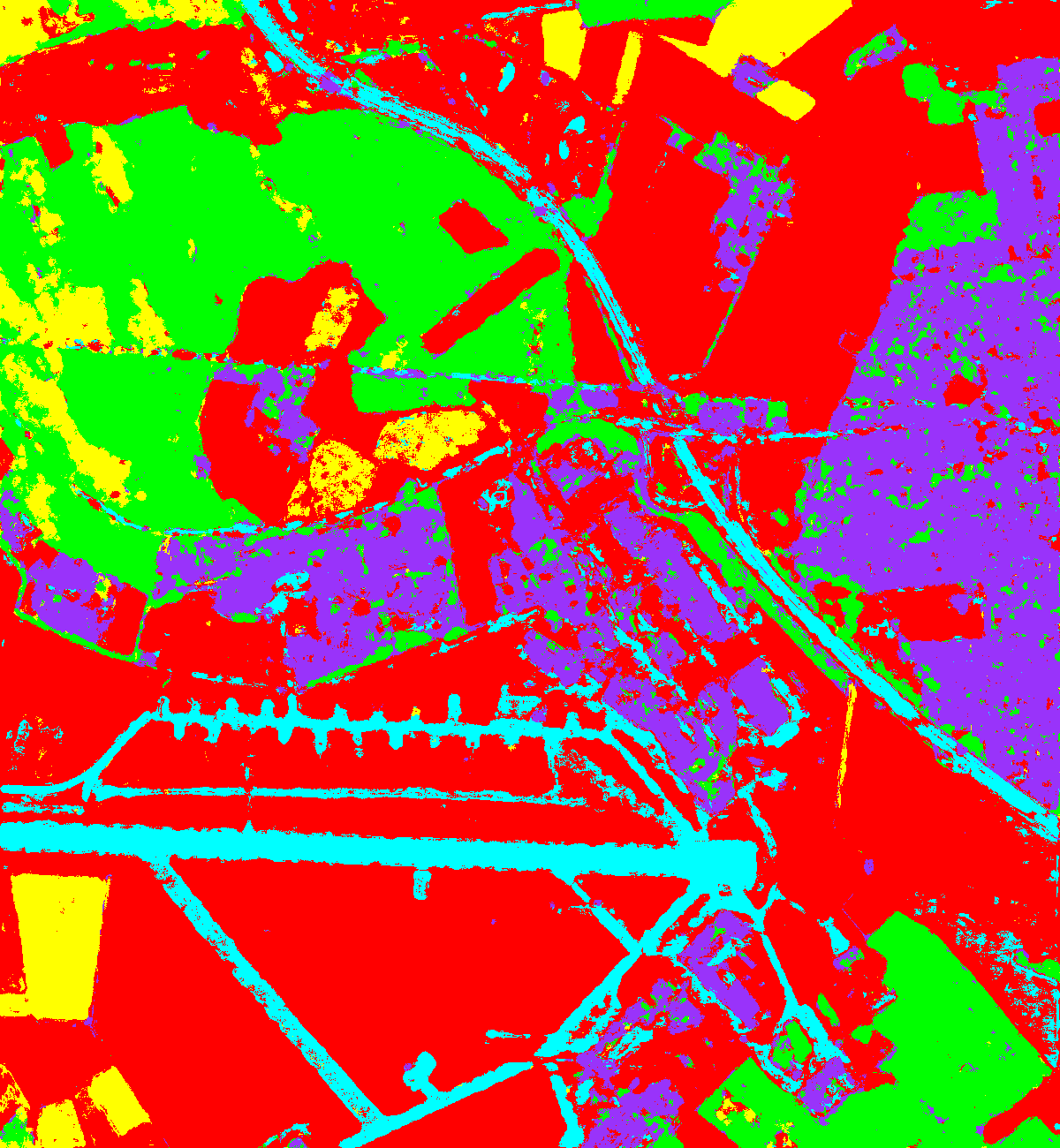}}}
	\subfigure[]{\fbox{\includegraphics[height=0.16\textheight]{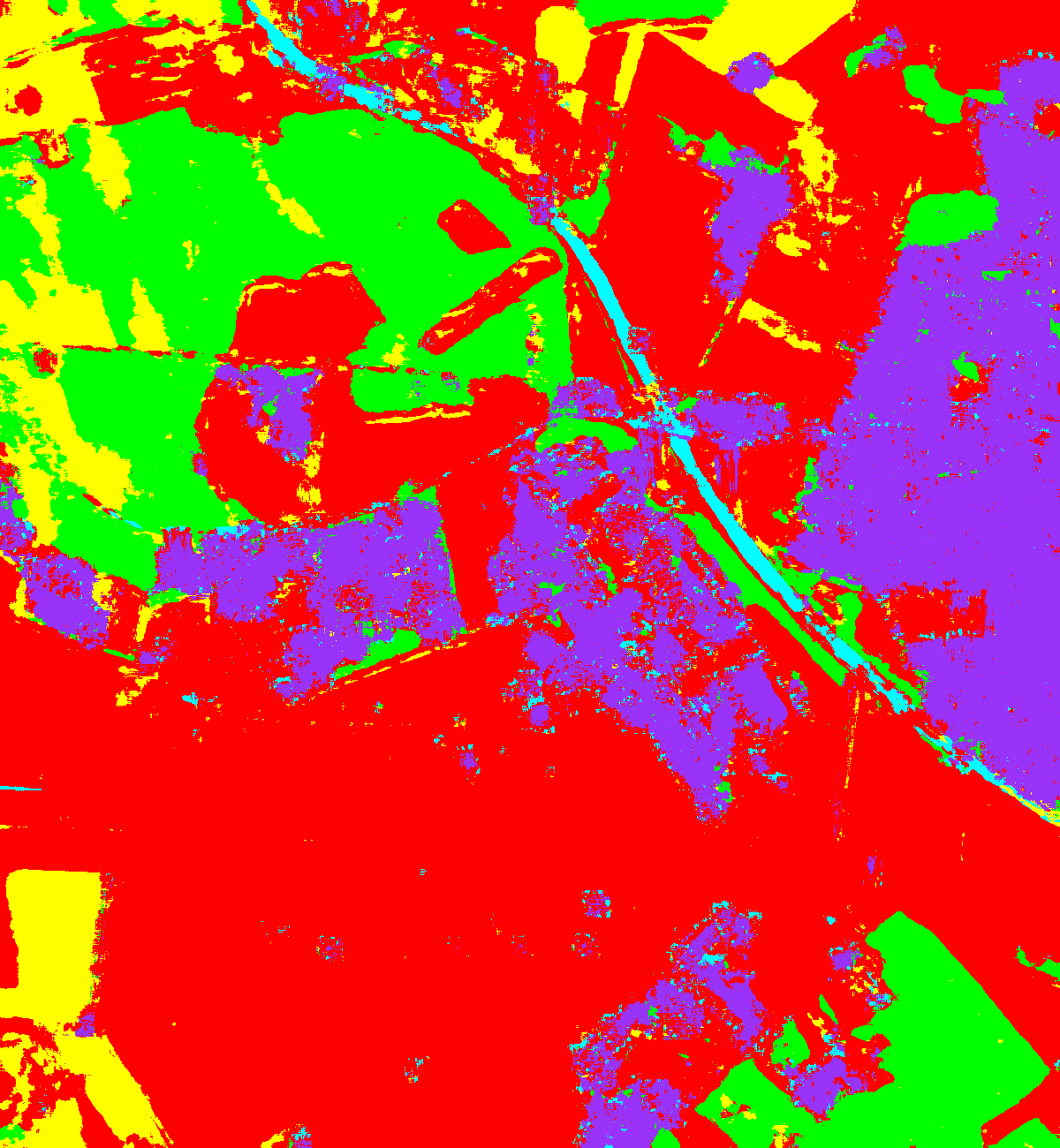}}}
          \subfigure[]{\fbox{\includegraphics[height=0.16\textheight]{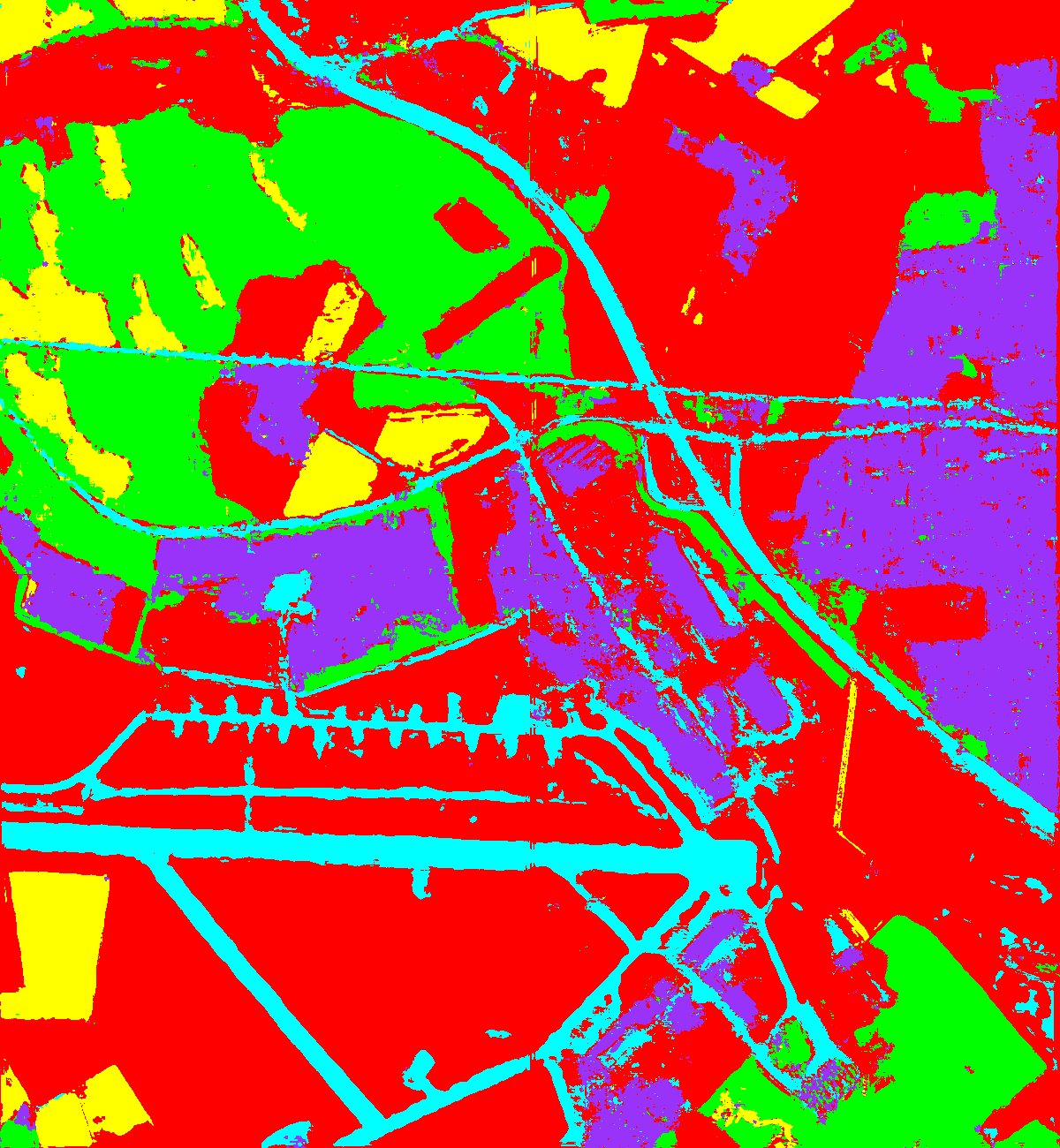}}}
	\subfigure[]
    {\fbox{\includegraphics[height=0.16\textheight]{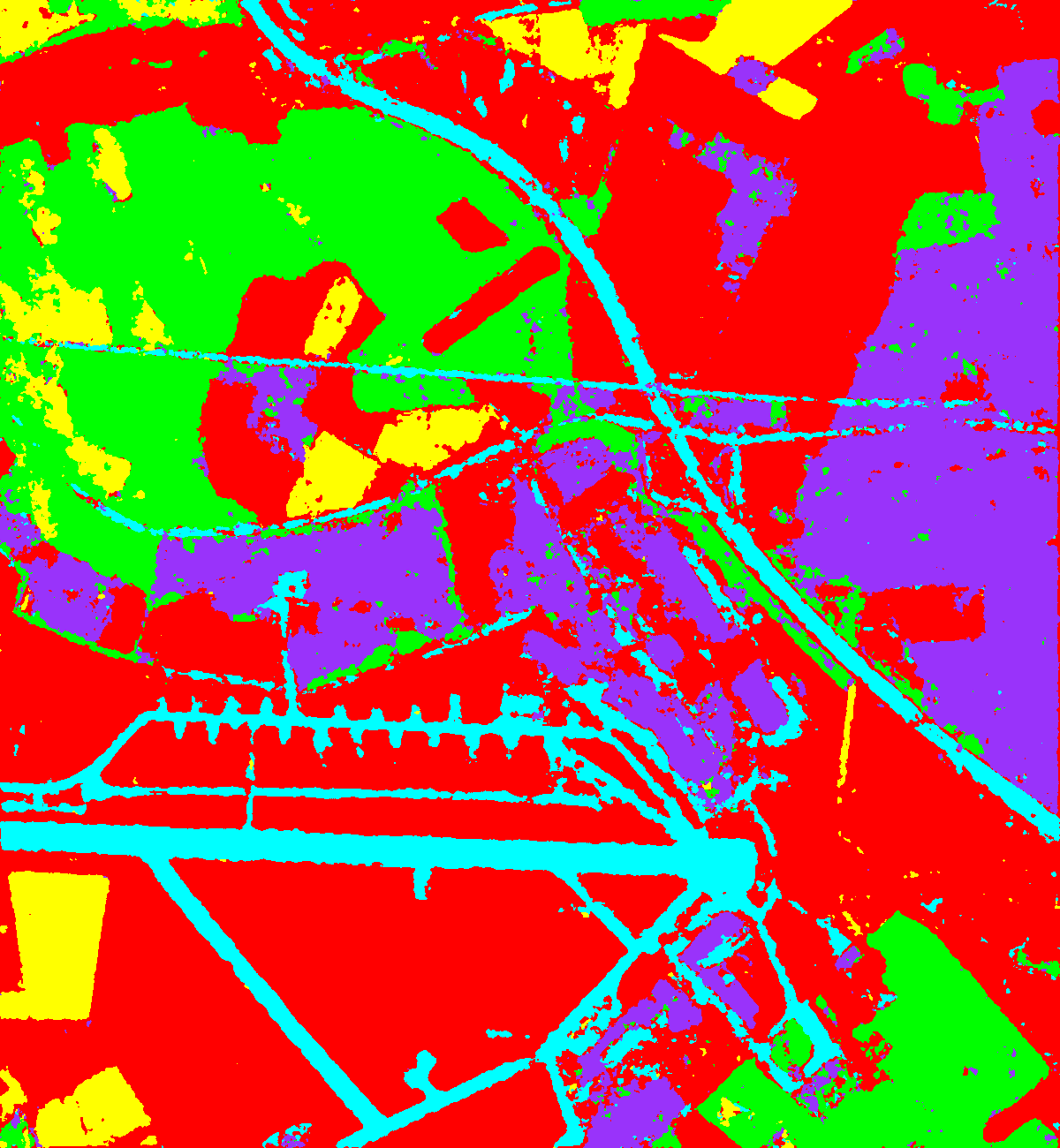}}}\\		
	\includegraphics[height=0.02\textheight]{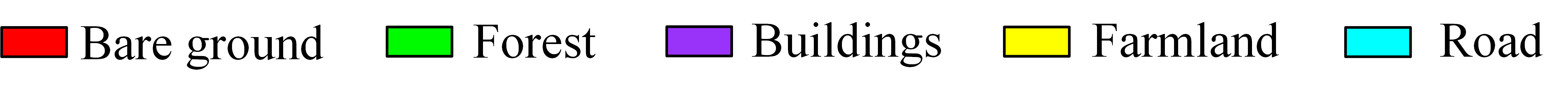}
	\caption{Classification results of different method on Oberpfaffenhofen area; (a)The label map; (b) CV-CNN;  (c) Super\_RF; (d) DFGCN; (e) AMS-MESL; (f) PolMPCNN; (g)HybridCVNet; (h)Proposed HPD\_CNN.}
	\label{f11}
\end{figure*}

 The classification results of different methods on the Oberpfaffenhofen dataset are presented in Figs.\ref{f11}(b)-(h), corresponding to CV-CNN, Super\_RF, DFGCN, AMS-MESL, PolMPCNN, HybridCVnet and HPD\_CNN, respectively. Figure \ref{f11}(a) is the label map for reference. As shown in Figure 2(b), the CV-CNN shows notable misclassification in the buildings category. The Super\_RF (Fig.\ref{f11}(c))method also appears obvious confusion between buildings and forest. The DFGCN in (d) can improve the classification performance, while some noisy points still exist in buildings class. The AMS-MESL also produces many noisy classes in both buidlings and farmland. The PolMPCNN in (e) improves the region homogeneity, while the road class is totally lost. The HybridCVnet performs well by leveraging complex-valued information, while it still has some noisy points. In contrast, the HPD\_CNN method (Fig.\ref{f11}(h)) can learn geometric structure of PolSAR data in manifold space to improve classification accuracy. Compared to the other methods (Fig.\ref{f11}(b)-(g)), the proposed method enhances both regional consistency and edge detail in the classification result.

Table \ref{t2} compares the classification accuracies of different methods on the Oberpfaffenhofen dataset. According to the table, it can be seen that the proposed algorithm gains the best classification accuracy compared to the other methods, with improvements in overall accuracy (OA) of 16.06\%, 9.40\%, 14.31\%, 8.26\%, 15.12\% and 2.29\% respectively. Specifically, the CV-CNN algorithm shows superior performance in road, while it cannot classify other terrain objects well with low accuracies, especially for bare ground class. The super\_RF method achieves high classification accuracy for the bare ground category, reaching 91.16\%, but the road category has a low classification accuracy, at only 56.05\%. The DFGCN also fails in the road class due to the small number of samples. The AMS-MESL method shows improved classification accuracy for the road category. The PolMPCNN almost totally lost the road which is consistent with the classification result in Fig.\ref{f11} (f). The hybridCVnet can obtain good classification performance by combining CV-CNN and CV-ViT. In contrast to these methods, the proposed algorithm achieves the highest classification accuracy for the road, buildings, and farmland categories. The overall accuracy (OA), average accuracy (AA), and Kappa coefficient of the proposed method are 90.94\%, 89.34\%, and 87.30\%, respectively, all outperforming the comparative algorithms.

\begin{tiny}
\begin{table*}
	\footnotesize
	\begin{center}
		\caption
		{ \label{t2}
			Classification accuracy of different methods on Oberpfaffenhofen Data Set(\%).}
		\begin{tabular}{lcccccccc}
			\hline
			class&CVCNN&Super\_RF&DFGCN&AMS-MESL&PolMPCNN&HybridCVnet&proposed\\
			\hline
			bare ground&68.86&\textbf{91.16}&89.37&89.93&86.92&\textbf{94.08}&93.36\\
			forest&81.16&66.44&86.47&85.75&82.51&90.22&\textbf{93.12}\\
			buildings&87.52&69.44&68.31&78.81&85.82&89.00&\textbf{89.54}\\
			farmland&70.49&87.90&37.70&68.51&65.65&\textbf{90.15}&88.21\\
			road&76.20&56.05&40.76&60.36&10.38&74.69&\textbf{80.47}\\
			OA&74.88&81.54&76.63&82.68&75.82&88.65&\textbf{90.94}\\
			AA&76.85&74.20&64.52&76.67&66.26&87.43&\textbf{89.34}\\
			Kappa&65.73&72.36&76.62&74.37&63.49&86.24&\textbf{87.30}\\
			\hline
		\end{tabular}
	\end{center}
\end{table*}
\end{tiny}

\subsection{Experimental results of Flevoland data set}

\begin{figure*}
	\centering
	\setlength{\fboxrule}{0.5pt}
	\setlength{\fboxsep}{0.01mm}

	\subfigure[]{\fbox{\includegraphics[height=0.16\textheight]{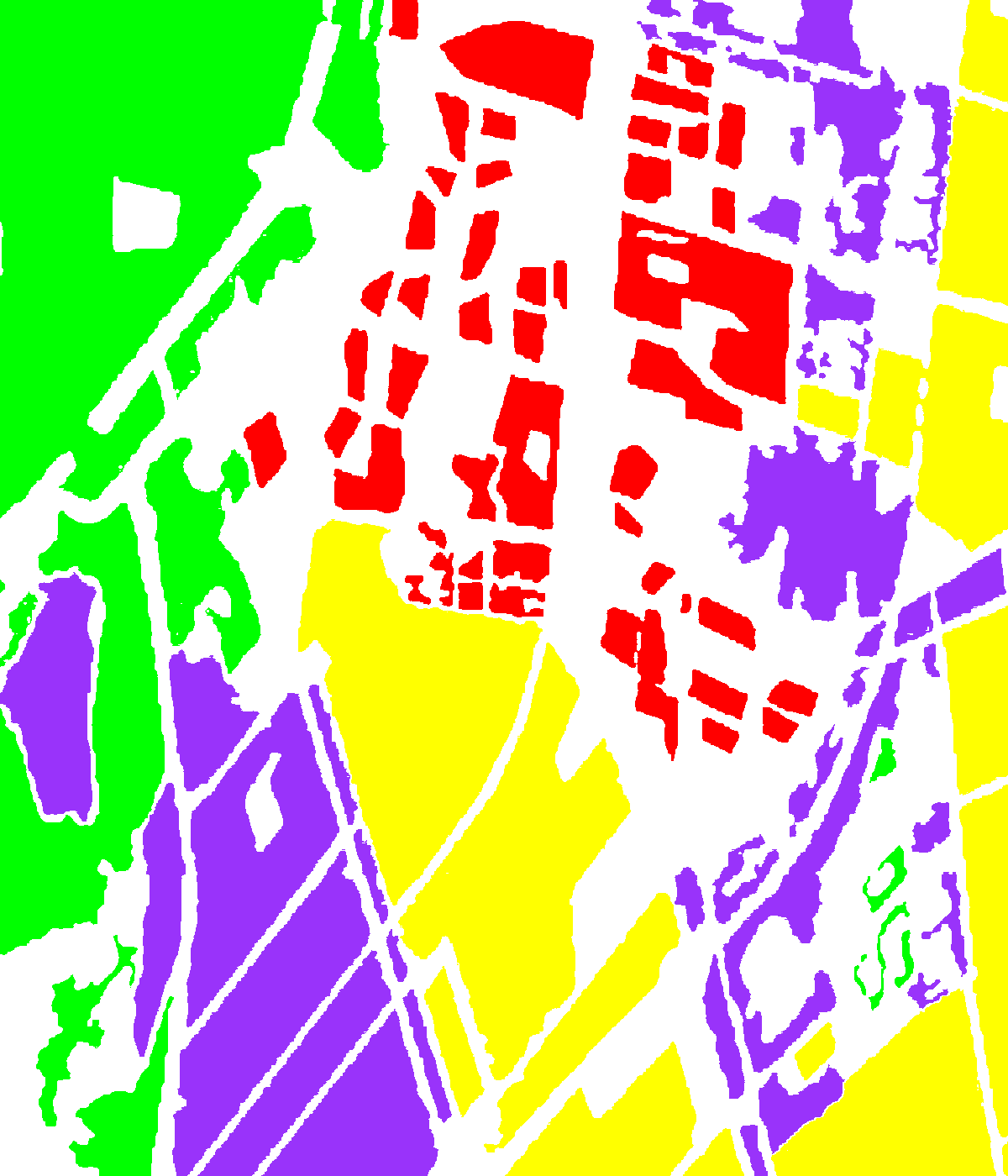}}}
	\subfigure[]{\fbox{\includegraphics[height=0.16\textheight]{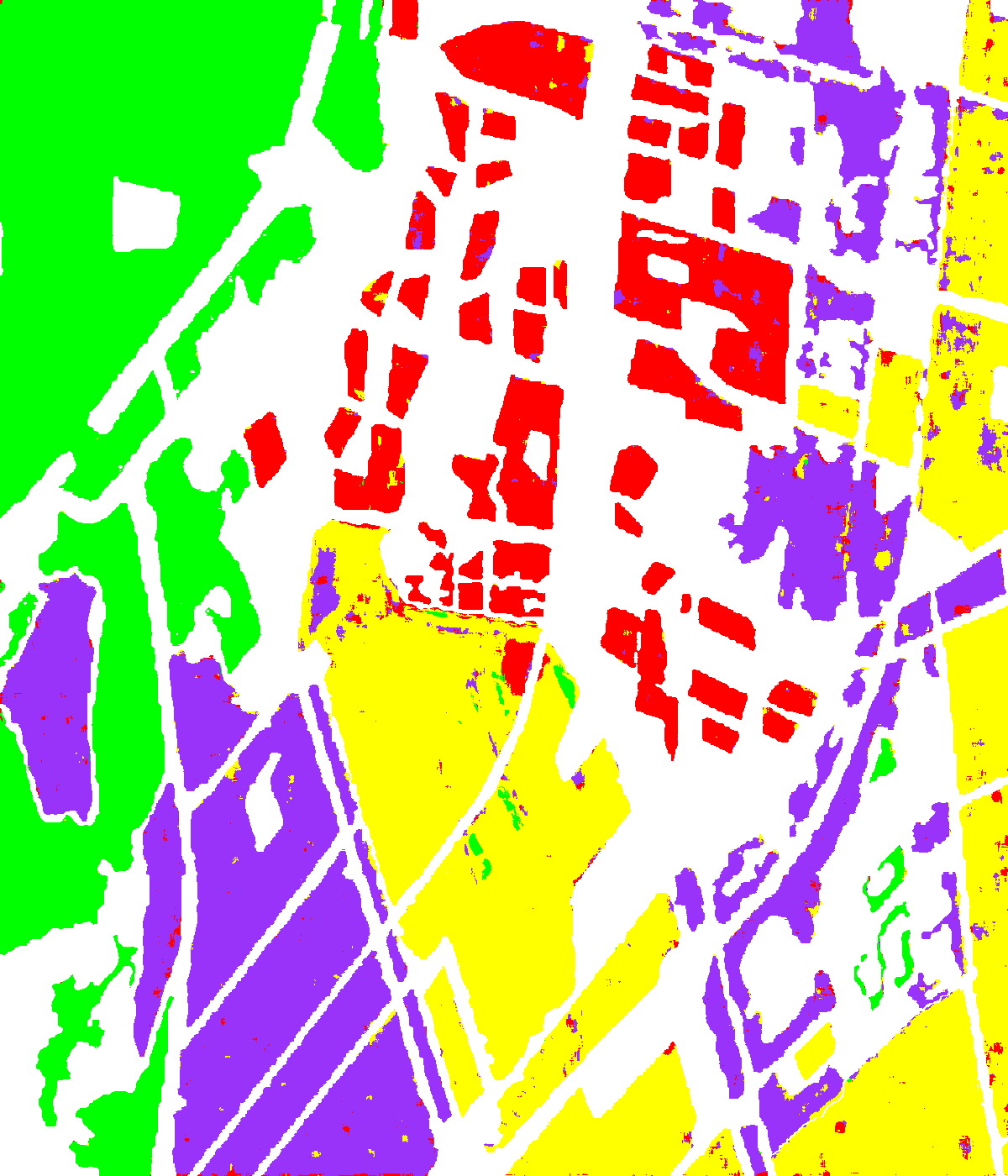}}}
	\subfigure[]{\fbox{\includegraphics[height=0.16\textheight]{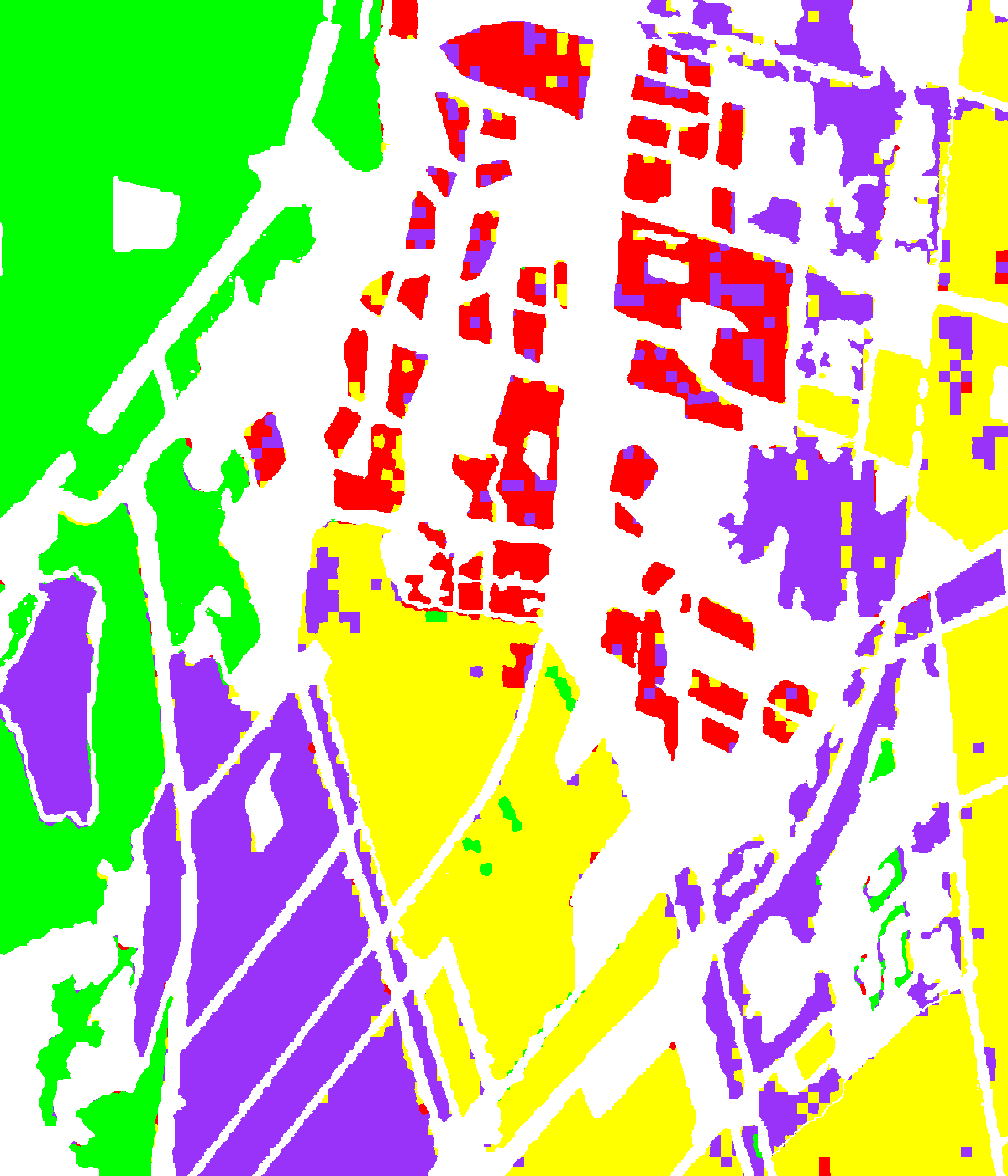}}}
	\subfigure[]{\fbox{\includegraphics[height=0.16\textheight]{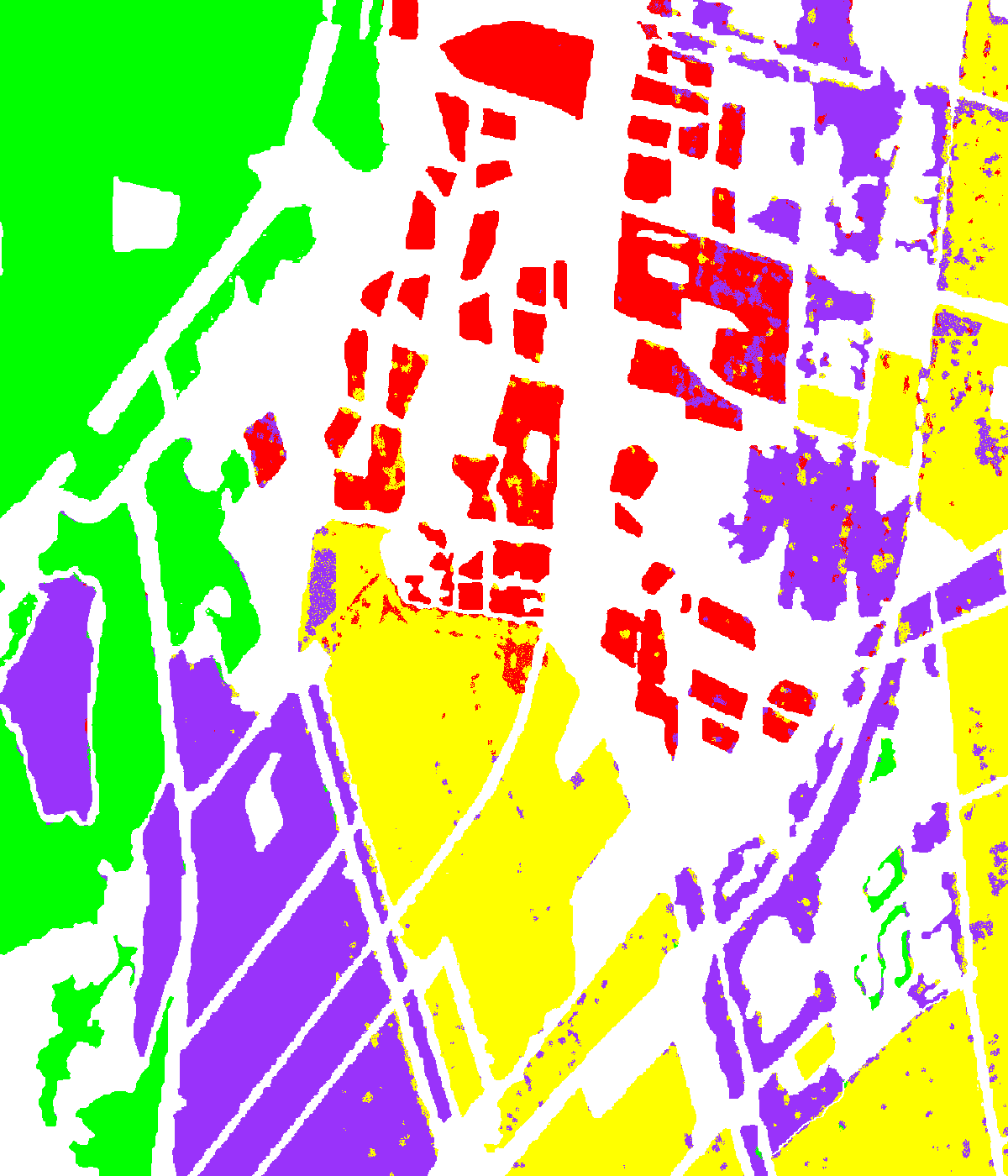}}}\\
	\subfigure[]{\fbox{\includegraphics[height=0.16\textheight]{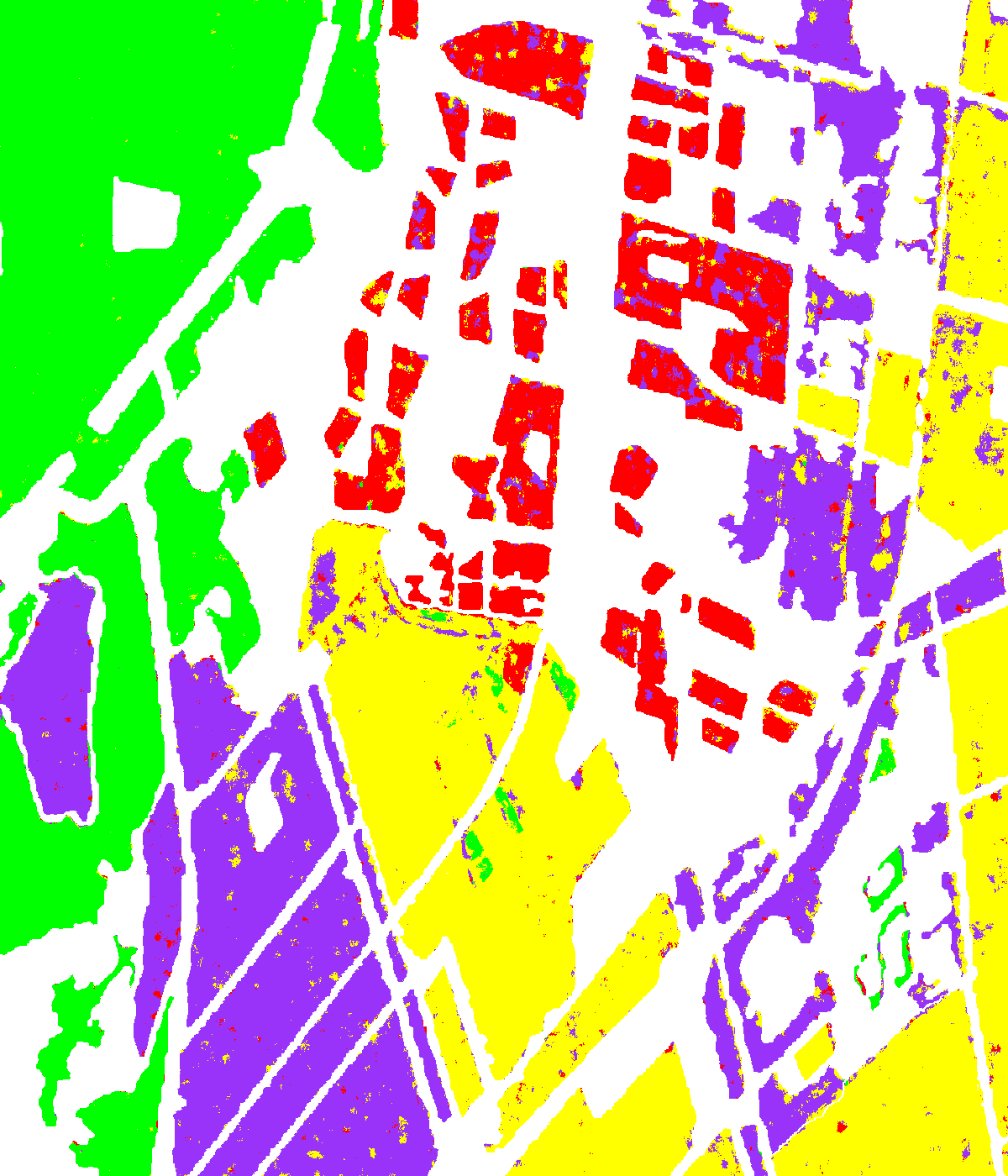}}}
	\subfigure[]{\fbox{\includegraphics[height=0.16\textheight]{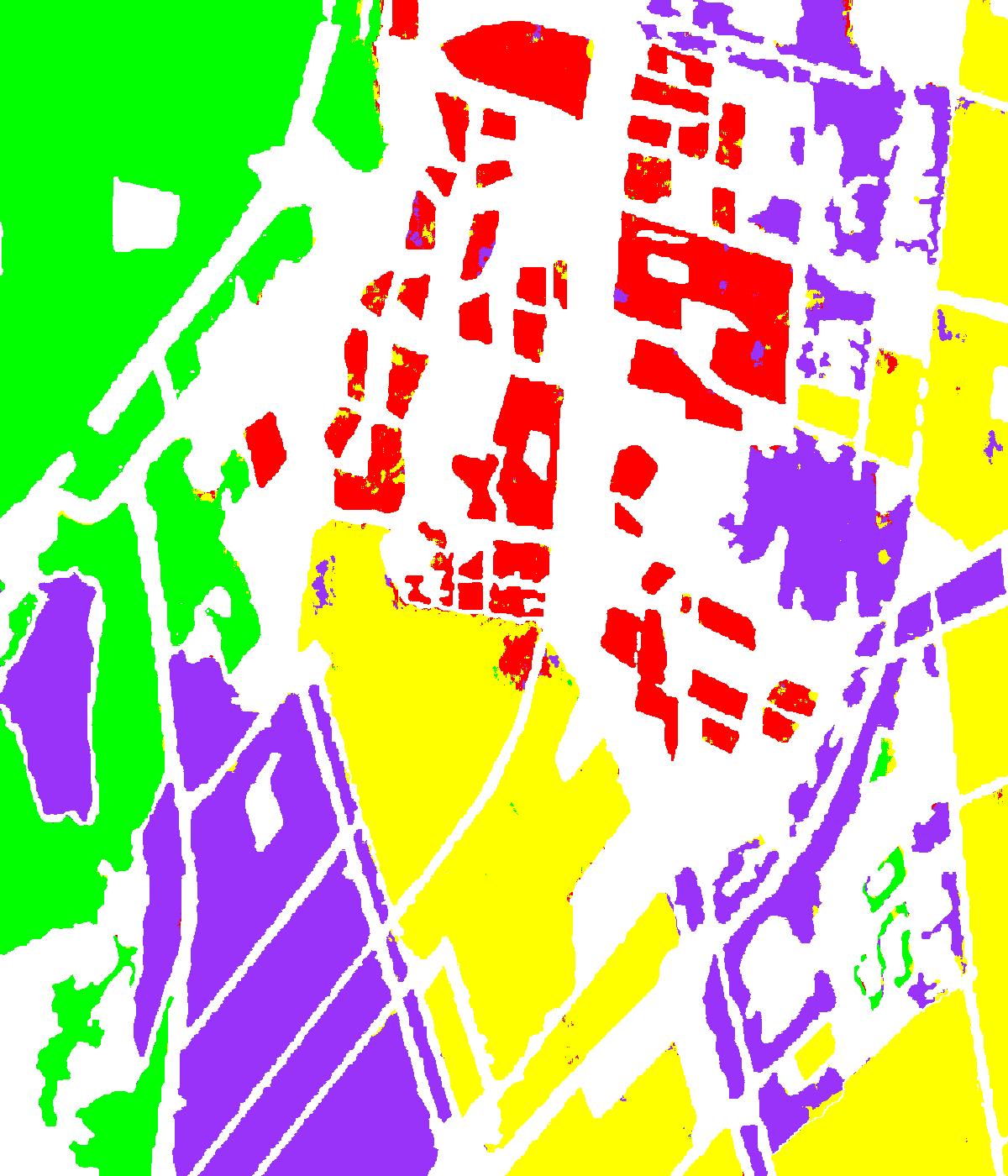}}}
          \subfigure[]{\fbox{\includegraphics[height=0.16\textheight]{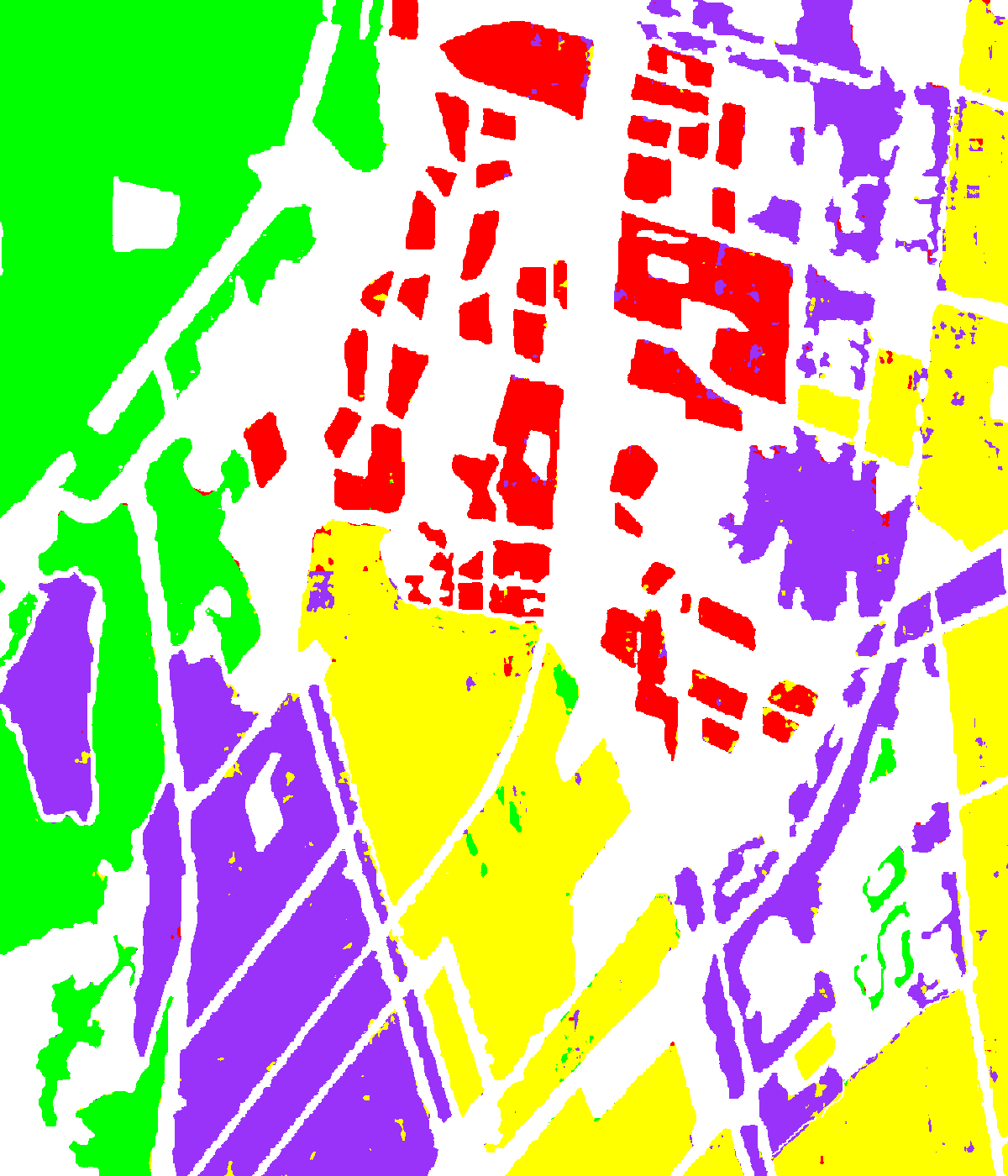}}}
	\subfigure[]
    {\fbox{\includegraphics[height=0.16\textheight]{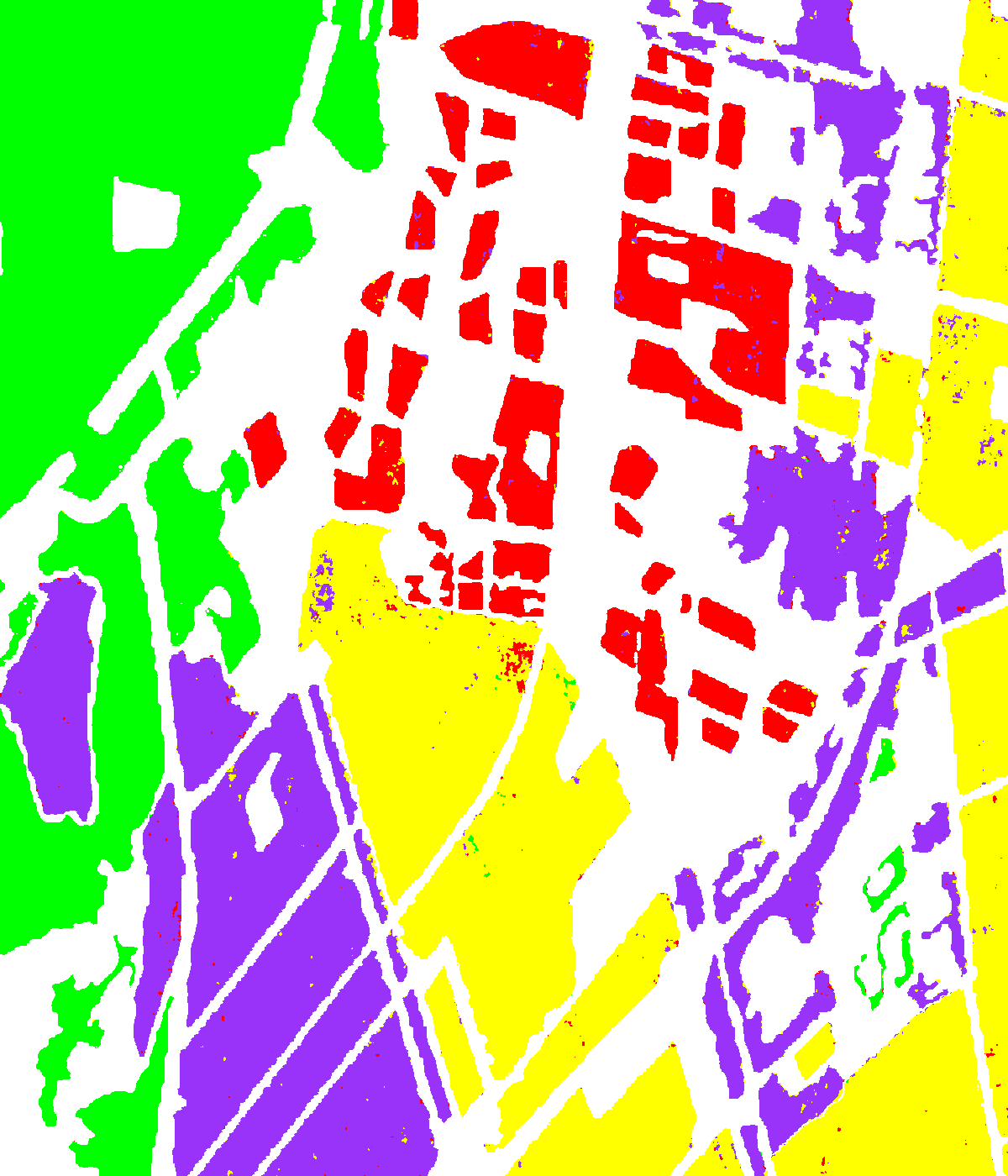}}}\\	
	\includegraphics[height=0.04\textheight]{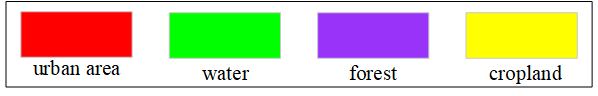}
	\caption{Classification results of different method on Flevoland area; (a)The label map; (b) CV-CNN;  (c) Super\_RF; (d) DFGCN; (e) AMS-MESL; (f) PolMPCNN; (g)HybridCVNet; (h)Proposed HPD\_CNN.}
	\label{f12}
\end{figure*}

The classification results of different methods are presented in Figs.\ref{f12}(b)-(h), corresponding to CV-CNN, Super\_RF, DFGCN, AMS-MESL, PolMPCNN, HybridCVnet and HPD\_CNN, respectively. Figure \ref{f12}(a) is the ground truth map as reference. As shown in Fig.\ref{f12}(b), the CV-CNN classification method shows many misclassifications in the cropland category, resulting in poor consistency. The super\_RF method in (c) leverages polarimetric information to improve classification performance, while an obvious block effect appears. The DFGCN also produces some noisy classes in cropland and urban area classes. The AMS-MESL method in (e) achieves better classification performance but shows notable misclassification in the urban area category. The PolMPCNN method improves the region homogeneity by removing noisy classes. HybridCVnet can obtain good performance due to complex information, while some misclassifications still exist in cropland. The proposed HPD\_CNN method achieves superior classification performance, with accurate details and fewer noise points in the classification map.

 Furthermore, Table \ref{t3} compares the classification accuracy of different methods in the Flevoland dataset. From the table, it is evident that the proposed HPD\_CNN method demonstrates exceptional classification performance compared to other compared methods, with improvements in OA of 2.16\%, 11.41\%, 3.04\%, 4.82\%, 0.64\% and 0.76\%, respectively. It shows a similar performance with PolMPCNN and HybridCVnet in this dataset, since this dateset is easily classified for various methods. However, the proposed method scores an obvious success in other datasets. Specifically, the CV-CNN method can perform well in urban and water classes, achieving accuracies of 96.26\% and 99.85\%, respectively. However, its classification accuracy for the cropland category is relatively low, at only 93.94\%. The Super\_RF method shows very low accuracy in both urban and woodland categories, with only 65.50\% and 84.39\%, respectively. The DFGCN has relatively low precision in urban and cropland, at 91.83\% and 93.82\%, respectively. The AMS-MESL method achieves high classification accuracy for the water category but performs extremely poorly in the urban category. The PolMPCNN and HybridCVnet methods perform well in various classes, while they are still slightly inferior to our method. The proposed HPD\_CNN method shows superior performance with an overall accuracy (OA) of 98.73\%, an average accuracy (AA) of 98.65\%, and a Kappa coefficient of 98.32\%.

\begin{table*}
	\footnotesize
	\begin{center}
		\caption
		{ \label{t3}
			Classification accuracy of different methods on Flevoland2 Data Set(\%).}
		\begin{tabular}{lcccccccc}
			\hline
			class&CVCNN&Super\_RF&DFGCN&AMS-MESL&PolMPCNN&HybridCVnet&proposed\\
			\hline
			urban&96.26&65.50&91.83&84.80&96.29&97.00&\textbf{98.00}\\
			water&99.85&92.87&99.57&99.19&99.14&99.68&\textbf{99.91}\\
			woodland&96.48&84.39&95.93&93.40&98.77&98.31&\textbf{98.23}\\
			cropland&93.94&94.90&93.82&93.85&98.66&97.17&\textbf{98.48}\\
			OA&96.57&87.32&95.69&93.91&98.09&97.97&\textbf{98.73}\\
			AA&96.63&84.42&95.29&92.81&97.42&97.68&\textbf{98.65}\\
			Kappa&95.33&52.20&94.11&91.66&97.14&97.51&\textbf{98.32}\\
			\hline
		\end{tabular}
	\end{center}
\end{table*}

\begin{figure}[ht]
	\centering
	\setlength{\fboxrule}{0.2pt}
	\setlength{\fboxsep}{0.01mm}
	\includegraphics[height=0.2\textheight]{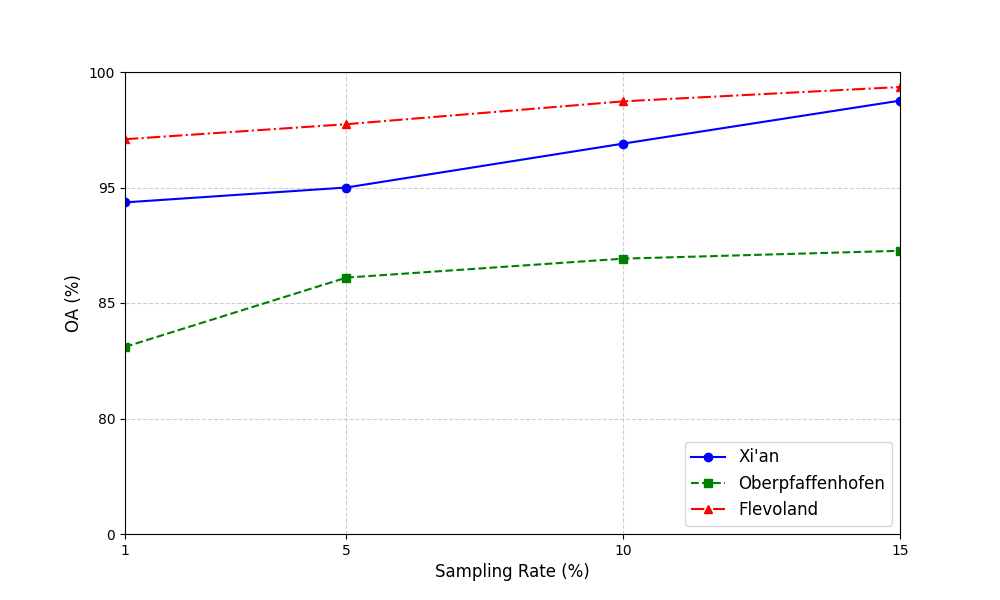}
	\caption{The effect of ratio of training samples on classification accuracy.}
	\label{fig14}
\end{figure}

\subsection{Ablation Study and Parameter Analysis}
Ablation studies are conducted on the Xi'an and Oberpfaffenhofen data sets to show the importance of each module in the proposed method. In addition, different parameters are discussed to show their effects on classification accuracy.

\textbf{ Ablation study on each module:} The proposed method comprises two essential modules: HPDnet and 3D-CVCNN enhanced modules. Here, we evaluate their importance and effects on the final result with OA and Kappa as evaluation metrics. The final classification accuracy is presented in Table \ref{t5}. It can be seen that the HPDnet with shallow features has a lower accuracy than the 3D-CVCNN model with high-level features. However, only the 3D-CVCNN is notably lower than the proposed HPD\_CNN model, indicating that HPD-based manifold learning is evidently effective for classification. The proposed HPD\_CNN method obtains superior performance by combining both advantages of the two modules.

\begin{table*}
\footnotesize
\begin{center}
\caption
{ \label{t5}
Classification accuracy of different modules on three Data sets(\%).}
\begin{tabular}{p{1.5cm}p{1.5cm}|p{1.5cm}p{1.5cm}|p{1.5cm}p{1.5cm}|p{1.5cm}p{1.5cm}} 
\hline
\multicolumn{2}{c|}{Modules}&\multicolumn{2}{c|}{Xi'an}&\multicolumn{2}{c|}{Oberpfaffenhofen}&\multicolumn{2}{c}{Flevoland}\\
\hline
HPDnet&3D-CVCNN&OA&Kappa&OA&Kappa&OA&Kappa\\
\hline
\checkmark&&80.53&68.49&71.35&55.90&85.58&80.25\\
&\checkmark&93.36&91.99&82.82&74.28&96.57&95.33\\
\checkmark&\checkmark&\textbf{96.90}&\textbf{95.56}&\textbf{90.94}&\textbf{87.30}&\textbf{98.73}&\textbf{98.32}\\
\hline
\end{tabular}
\end{center}
\end{table*}


\textbf{Effect of the ratio of training samples:}
The ratio of training samples is a crucial parameter for deep learning. In this study, we tested the effect of the sample ratio on the classification performance in three data sets. The sample ratio varies from 1\% to 15\% with an interval of 5\%, as shown in Fig.\ref{fig14}. It is observed that there is a significant improvement in all classification metrics from 1\% to 10\%. The sample ratio with 10\% can achieve excellent performance. In addition, too many training samples may increase the computation time. Therefore, we select 10\% as the ratio of training samples.

\textbf{Effect of the patch size on classification performance:}
Patch size is a crucial parameter of the proposed HPD\_CNN model. A small patch can greatly reduce computing time, while an extremely small patch may lose semantic information. In this experiment, we test different patch sizes with $9\times 9$, $13\times 13$, $17\times 17$,respectively. The effect of different patch sizes on the classification accuracy can be shown in Table \ref{t4} in Xi'an dataset. According to the table, we can see that the OA values grow rapidly from $9\times 9$ to $13\times 13$, while $13\times 13$ can achieve a similar accuracy with less computing cost than $17\times 17$. Therefore, we select the patch size as $13\times 13$.
\begin{table}
\footnotesize
\begin{center}
\caption
{ \label{t4}
 The effect of different patch size on Xi'an dataset(\%).}
\begin{tabular}{p{1.5cm}p{1cm}p{1cm}p{1cm}}
\hline
patch size&$9\times 9$&$13\times 13$&$17\times 17$\\
\hline
OA&93.69&96.90&97.30\\
AA&92.56&96.11&96.28\\
Kappa&93.67&95.56&96.15\\
\hline
\end{tabular}
\end{center}
\end{table}

\textbf{Analysis of running time:}
Xi'an data set is utilized to analyze the running time of the compared and proposed methods. Table \ref{t6} presents the training and testing time. In addition, to test the computation effectiveness of the proposed CM\_ASQRT acceleration module, we compare the running time between the proposed method without and with the CM\_ASQRT module, noted by "HPD\_CNN-o" and "HPD\_CNN", respectively. It can be seen that the proposed method with CM\_ASQRT module ("HPD\_CNN") can greatly reduce training time compared to the "HPD\_CNN-o", since the eigenvalue decomposition is a time-costing operation with the solution of matrix's inverse and trace. Besides, the PolMPCNN exhibits the longest training and test times, which can be attributed to its input feature dimension and large-scale convolution. However, the Super\_RF demonstrates the shortest training time without deep learning. The proposed method can obtain the best classification performance in a relatively short time, demonstrating the effectiveness of the proposed method in terms of both time efficiency and performance.

\begin{table*}
	\footnotesize
	\begin{center}
		\caption
		{ \label{t6}
			Running time of different methods on Xi'an Data Set ($s$)}
		\begin{tabular}{lcccccccc}
			\hline
			time&CVCNN&super\_RF&DFGCN&AMS-MESL&PolMPCNN&HybridCVnet&HPD\_CNN-o&HPD\_CNN\\
			\hline
			train&3463.20&59.22&475.62&345.50&26100.35&1023.21&663.24&152.35\\
			test&38.43&1.85&7.85&3.12&327.53&80.20&37.82&7.98\\
			\hline
		\end{tabular}
	\end{center}
\end{table*}

\section{Conclusion}

This paper has presented a novel Riemannian-to-Euclidean complex HPD convolution network(HPD\_CNN) for PolSAR image classification. This is the first time a complex HPD manifold network has been proposed for PolSAR images in Riemannian space. The proposed method consists of two modules: a complex HPD unfolding network in Riemannian space and a CV-3DCNN network in Euclidean space. Firstly, a complex HPD unfolding network is defined in Riemannian space, which constructs a set of HPD mapping and rectify operations, similar to convolution and Relu, then the ReEig operation is designed to covert the manifold matrix to Euclidean space. The proposed HPDnet effectively preserves geometric structures by ensuring that all operations are performed within the manifold space. Secondly, a CV-3DCNN network is followed to learn contextual information to reduce speckle noises and enhance feature representation. Experiments demonstrate that the proposed method can achieve superior quantitative and qualitative results compared to state-of-the-art methods.

In addition, the HPDnet is a matrix-based network that can learn geometric structure of matrices, while it ignores the contextual information of PolSAR images. In the further work, we will attempt to exploit how to directly learn the contextual information of complex matrices in manifold space for PolSAR images.

\section*{Acknowledgments}

This work was supported in part by the National Natural Science Foundation of China under Grant 62471387,62272383,62372369, in part by the Youth Innovation Team Research Program Project of Education Department in Shaanxi Province under Grant 23JP111.

\bibliographystyle{IEEEtran}
\bibliography{mybibfile}

\end{document}